\documentclass[lettersize,journal]{IEEEtran}
\usepackage{amsmath,amsfonts}
\usepackage{algorithmic}
\usepackage{algorithm}
\usepackage{array}
\usepackage[caption=false,font=normalsize,labelfont=sf,textfont=sf]{subfig}
\usepackage{textcomp}
\usepackage{stfloats}
\usepackage{url}
\usepackage{verbatim}
\usepackage{graphicx}
\usepackage{tikz}
\usepackage{cite}
\usepackage{siunitx}
\usetikzlibrary{arrows.meta}
\newcommand{\overlaysubfig}[3][]{
\begin{tikzpicture}
\node[anchor=south west,inner sep=0] (image) at (0,0) {\includegraphics[#1]{#3}};
\node[anchor=north west,fill=white,fill opacity=0.85,text opacity=1,inner sep=1pt,font=\scriptsize\sffamily\bfseries] at ([xshift=1mm,yshift=2mm]image.north west) {(#2)};
\end{tikzpicture}
}
\usepackage[whole]{bxcjkjatype}
\usepackage{tabularray}
\usepackage{booktabs}
\usepackage{multirow}
\usepackage{multicol}
\usepackage{amssymb} 
\usepackage{pifont}  
\usepackage{threeparttable}

\newcommand{\cmark}{\checkmark}
\newcommand{\xmark}{--}
\newif\ifjp
\jpfalse

\begin{document}

\title{
SHAPE: Simultaneous Water Hydraulic Actuation\\ and Position Estimation of a Sensorless Remote Actuator\\ through a Thin and Long Flexible Tube
}

\author{Yuki~Nakamura,~\IEEEmembership{Student~Member,~IEEE}, Shuto~Yoshimura,~\IEEEmembership{Student~Member,~IEEE},\\Tomoyuki~Noda,~\IEEEmembership{Member,~IEEE}, and Yoshihiro~Nakata,~\IEEEmembership{Member,~IEEE}

\thanks{This work was partially supported by JSPS KAKENHI (Grant Number JP24K21325) and conducted in collaboration with Chugai Technos Corporation.
\textit{(Yuki Nakamura and Shuto Yoshimura are co-first authors.) (Corresponding author: Yuki Nakamura.)}
}

\thanks{Y. Nakamura is with the Department of Mechanical and Intelligent Systems Engineering, Graduate School of Informatics and Engineering, The University of Electro-Communications, 1-5-1 Chofugaoka, Chofu, Tokyo, 182-8585, Japan (e-mail: yuki.nakamura@uec.ac.jp), and also with Chugai Technos Corporation, 4-3-1 Ishiuchikita, Saeki-ku, Hiroshima, 731-5109, Japan.}

\thanks{S. Yoshimura and Y. Nakata are with the Department of Mechanical and Intelligent Systems Engineering, Graduate School of Informatics and Engineering, The University of Electro-Communications, 1-5-1 Chofugaoka, Chofu, Tokyo, 182-8585, Japan (e-mail: y2432118@edu.cc.uec.ac.jp; ynakata@uec.ac.jp).}

\thanks{T. Noda is with the Department of Brain Robot Interface, ATR Computational Neuroscience Laboratories, 2-2-2 Hikaridai Seika-cho, Soraku-gun, Kyoto, 619-0288, Japan (e-mail: t\_noda@atr.jp) and also with the Department of Mechanical and Intelligent Systems Engineering, Graduate School of Informatics and Engineering, The University of Electro-Communications, 1-5-1 Chofugaoka, Chofu, Tokyo, 182-8585, Japan.}
}

\markboth{Journal of \LaTeX\ Class Files,~Vol.~xx, No.xx, mm~xxxx}
{Nakamura \MakeLowercase{\textit{et al.}}:SHAPE: Simultaneous Water Hydraulic Actuation and Position.
Estimation of Sensorless Remote Actuators Using a Thin
Extra-Length Tube}

\maketitle

\begin{abstract}
Robot sensors and electronic equipment are prone to failure in harsh environments.
With water hydraulic drive, thin and long tubes enable remote operation without actuator-side sensors.
Furthermore, the elasticity of the tubes reduces the impedance of the joints (actuators), benefiting robot tasks involving unexpected contact with the environment or vibrations.
However, owing to the low impedance and limited camera visibility, accurately positioning the joint (or end effector) to the target location under varying load conditions is challenging.
This study proposes a novel method that employs water-filled flexible tubes to enable the transmission of driving power and actuator-side information to and from the actuator, respectively, without actuator-side sensors.
By modeling volumetric loss during transmission based on pressure fluctuations and incorporating minor air entrapment, simultaneous power transmission and position estimation is achieved through a tube up to 50 m.
Thus, it becomes possible to use a feedback control framework that was previously difficult to implement in sensorless systems.
Experimental validation confirms stable position control of a sensorless water hydraulic cylinder under varying loads.
Furthermore, a field parameter-identification method accounts for tube and air entrainment variability without requiring actuator-side sensors.
These contributions promote reliable remote control of robots in harsh environments.
\end{abstract}


\section{Introduction}
In harsh environments, such as radiation-exposed and underwater areas, robot sensors and electronic equipment are prone to failure and malfunction~\cite{sharp1996radiation}, making robot construction difficult.
Water hydraulic drive is promising because it allows for sensorless remote operation by using long and thin tubes ~\cite{nakamura2025remotehydraulic,Fukumoto2018} and offers a high torque-to-weight ratio~\cite{suzumori2020new}.
However, while these robots are often operated based on external visual information~\cite{kobayashi2023discussion}, without actuator-side sensors, the actuator position cannot be determined in confined spaces, limited bandwidth, or poor visibility.

In damaged plants and disaster sites, poor visibility and the risk of collision make reliable positioning at low speeds more important than at high speeds.
Even at low speeds, when the load fluctuates, volume loss occurs due to the expansion of the flexible tube, the pseudo-compressibility of water, and air entrapment.
Therefore, the actuator position cannot be directly determined from the displacement of the supply source by relying on the incompressibility of water.
Prior studies have mainly addressed transmission dynamics or delay compensation in hydraulic pipes~\cite{ganesh2004dynamics,guo2018compact,simonelli2019hydrostatic,dong2019high,Bolignari2020}; however, kinematic feedback generally requires local sensors, and providing somatosensory information, particularly position, remains unresolved.

\begin{figure}[tbp]
\centering
  \includegraphics[width=\linewidth]{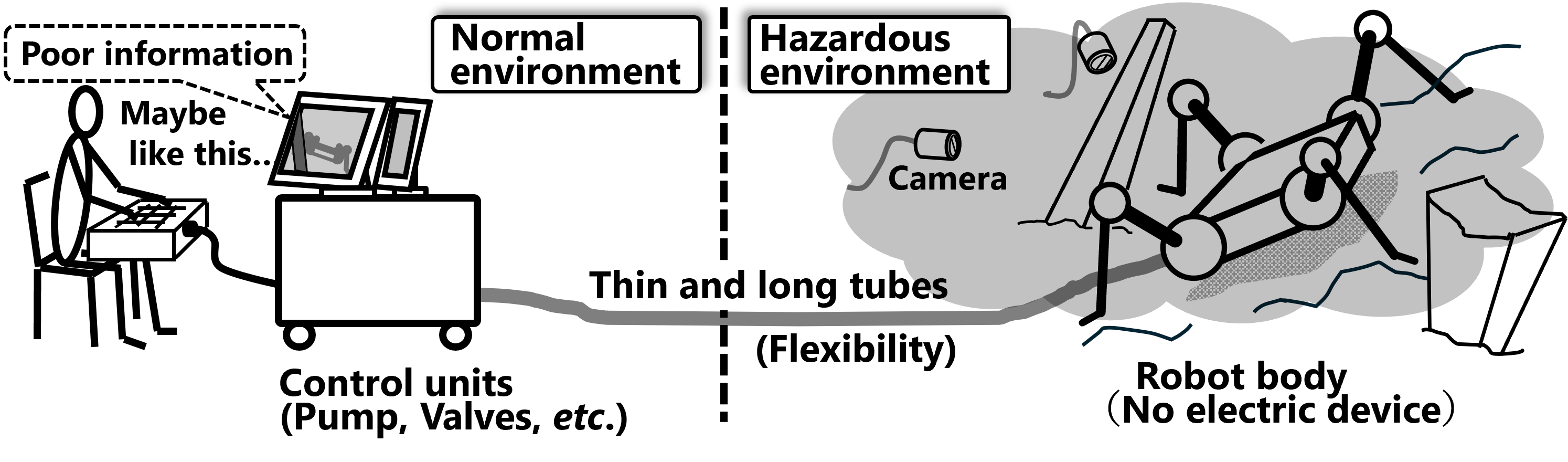}

  \caption{Sensorless remote water hydraulic drive robot system. The actuator is driven through a thin and long water-filled tube.
  }
\label{remote_robot_system}       
\end{figure}

This study addresses the sensorless remote actuation problem, as shown in Fig.~\ref{remote_robot_system}.
Simultaneous Water Hydraulic Actuation and Position Estimation (SHAPE) is defined as the simultaneous transmission of driving power and actuator-position information through a single thin and long water-filled flexible tube (Fig.~\ref{Power_and_information}).
In SHAPE, the tube serves as a power conduit and an information channel for inferring the kinematic state of the remote actuator.
Under quasi-static operation, pressure-dependent tube expansion and fluid pseudo-compressibility cause volumetric loss between the source and actuator.
By modeling this loss, the actuator position can be estimated from the displacement and pressure measurements on the source side, enabling position control without actuator-side sensors.

This study makes the following key contributions:
\begin{itemize} 
\item SHAPE framework: Simultaneous transmission of driving power and position information is achieved through a single thin and long water-filled tube, enabling feedback control of a remote sensorless actuator.
\item Volumetric-loss model: Pressure-induced tube expansion and fluid pseudo-compressibility, including minor air entrapment, are modeled to estimate the distal actuator position from source-side measurements.
\item Displacement-based Flow Measurement Cylinder (DFMC) and sensorless in-situ parameter identification: The DFMC is introduced to demonstrate that tube stiffness and air-volume parameters can be identified under sensorless actuator-side conditions.
\item Experimental validation: Stable sensorless position control of a water hydraulic cylinder is demonstrated through a 50~m tube under varying load conditions using a basic Proportional-Integral-Derivative (PID) controller.
\end{itemize}

\section{Comparison with Previous Research}
\subsection{Conceptual framework}

\begin{figure}[tbp]
\centering
    \subfloat{
        \overlaysubfig[clip,width=0.9\linewidth]{a}{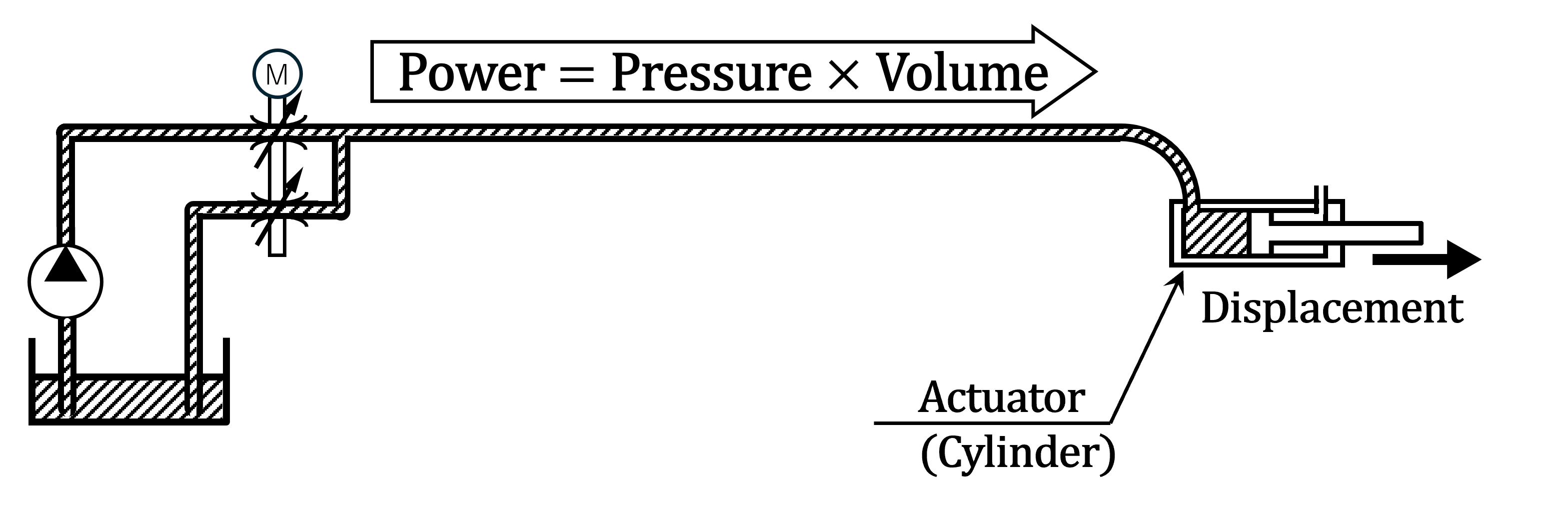}
    }
    \\
    \subfloat{
        \overlaysubfig[clip,width=0.9\linewidth]{b}{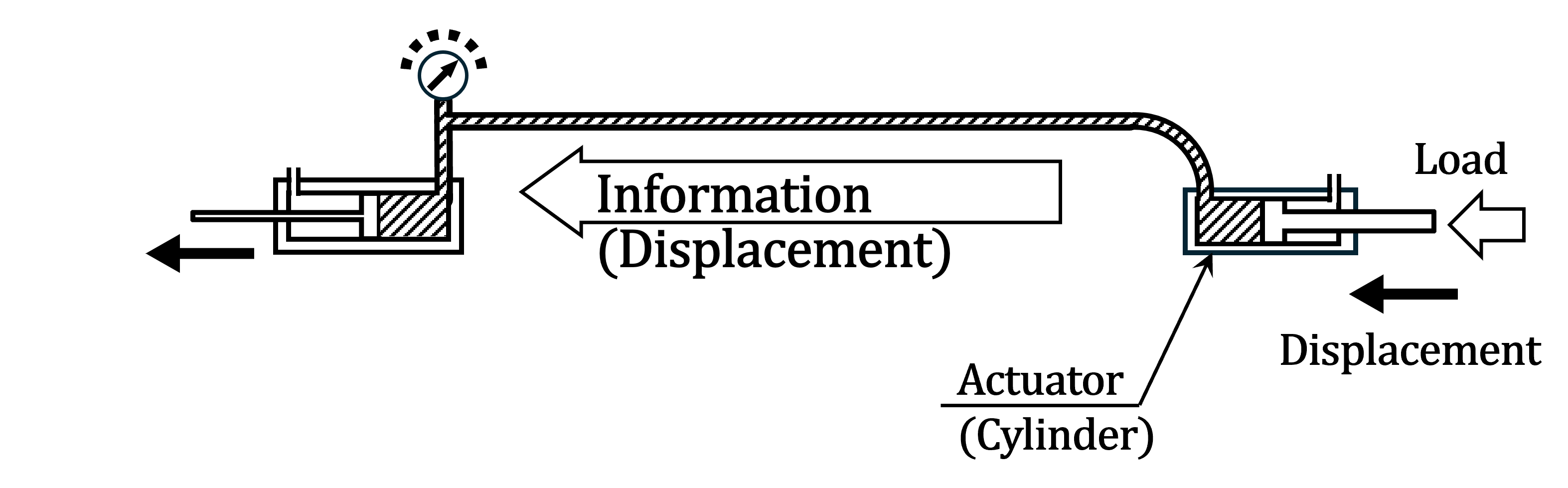}
    }
    \\
    \subfloat{
        
        \overlaysubfig[clip,width=0.9\linewidth]{c}{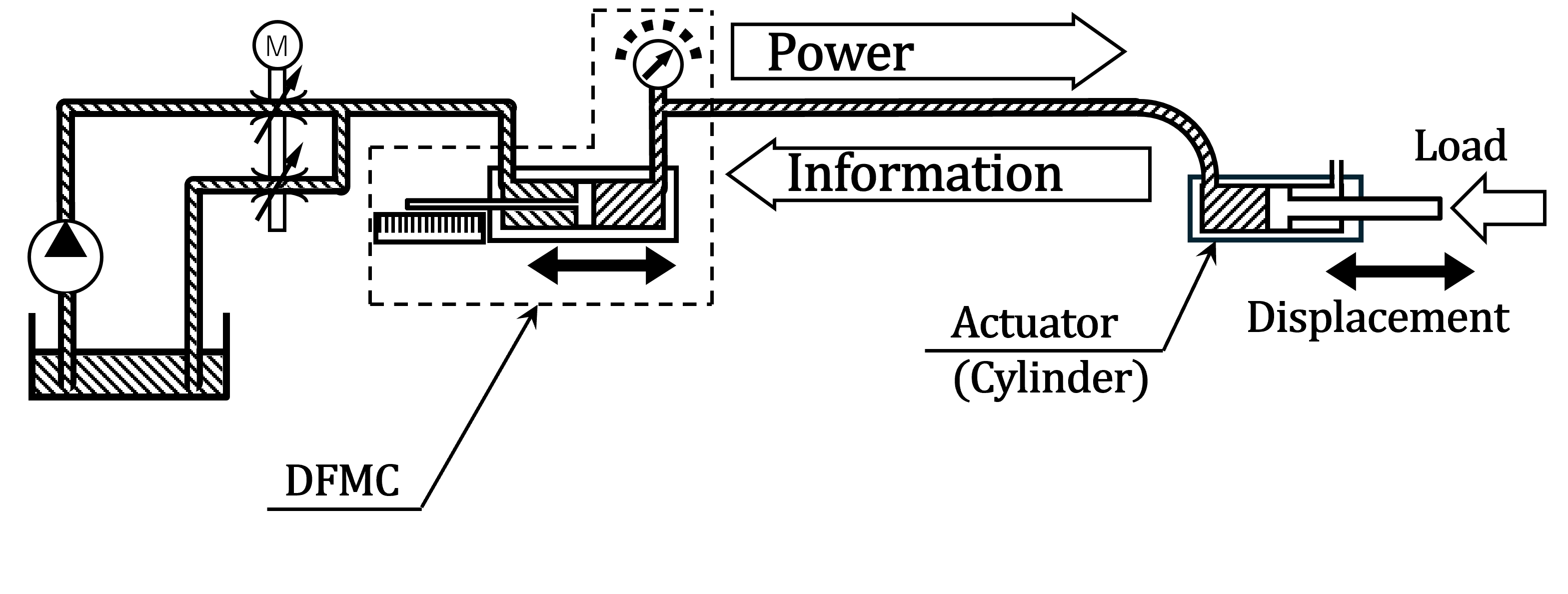}
    }
    \\

\caption{Power and information transmission through the water-filled tube. (a) Driving power to the actuator. (b) When two cylinders are connected, the displacement of the right cylinder can be estimated by measuring the displacement of the left cylinder. (c) Illustration of the SHAPE problem and approach with the DFMC.}
\label{Power_and_information}
\end{figure}

As water has low kinematic viscosity and high stiffness, it can transmit hydraulic power through a long and thin tube while also conveying pressure-related information from the distal side. Rather than assuming exact equality between source- and actuator-side pressures, SHAPE exploits rapid pressure-related information propagation and pressure-dependent volume change to transmit power and information through a long and flexible water-filled tube (Fig.~\ref{Power_and_information}(a)(b)).
As shown in Fig.~\ref{Power_and_information}(c), a source-side DFMC drives the sensorless actuator through a long and thin tube and measures the source-side pressure and displacement. A physical model then compensates for the volumetric loss caused by tube expansion and fluid pseudo-compressibility.
Furthermore, when a non-moving actuator state can be assumed, the pressure and displacement of the DFMC correspond to the relationship between the system pressure and the volume loss in the transmission. This relationship enables on-site parameter re-identification without requiring additional sensors on the actuator side.

\begin{figure}[tbp]
\centering
    \subfloat{
        \overlaysubfig[width=0.9\linewidth]{a}{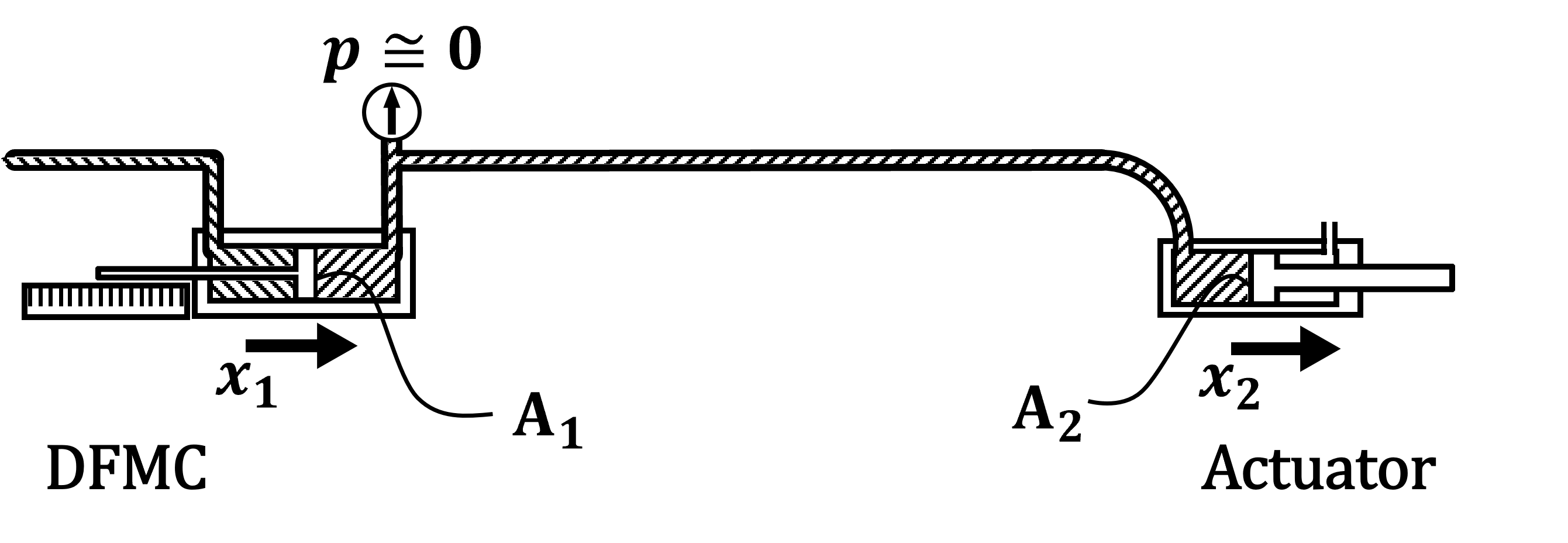}
    }\\
    \subfloat{
        \overlaysubfig[width=0.9\linewidth]{b}{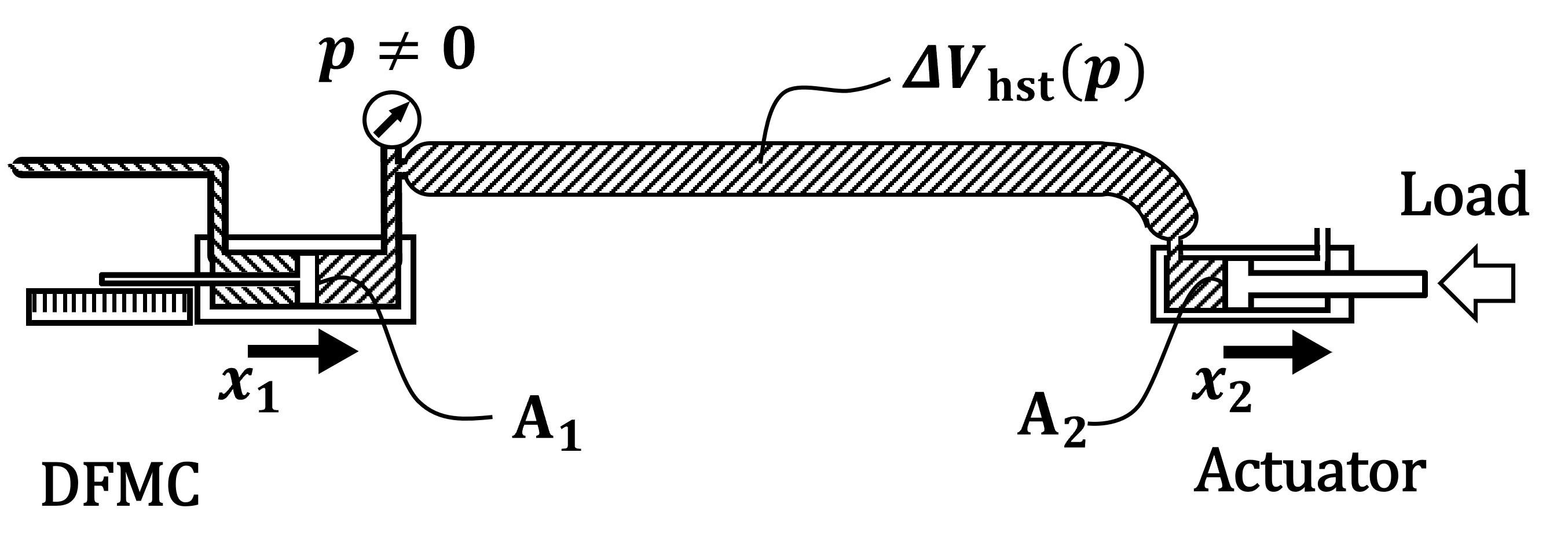}
    }
\caption{Approach to solving the SHAPE problem. (a) No load condition. (b) With load condition.}
\label{Base idea}
\end{figure}

\subsection{Previous research}

Table~\ref{table_comparison_study} presents a comparison with previous studies on remote driving of water hydraulic actuators. 
To the best of our knowledge, prior studies have not explicitly addressed the variation in the hydraulic system stiffness caused by changes in the actuator load during operation.

Several studies have explored the use of Hydrostatic Transmission (HST) to drive actuators without a sensor, particularly in systems designed for operation within magnetic resonance imaging (MRI) devices~
\cite{ganesh2004dynamics,guo2018compact,simonelli2019hydrostatic,dong2019high}. 
Sensorless systems are required because sensors adversely affect MRI-generated images.
The operator remotely controls the leader--follower system using HST.
Previous studies have reported remote distances of up to approximately 10 m, with the practitioner operating the leader device while visually checking the inspection area or using MRI images.
To improve transmission performance between leaders and followers, researchers have sought to increase the rigidity of the HST section.
Consequently, the use of highly rigid fluids, such as water and rigid piping, is desirable.
However, certain applications require pipe flexibility to accommodate bundling or to achieve compact device configurations.
In piping design, rigidity and flexibility represent a trade-off.

In studies on systems designed for MRI environments, stiffness has been evaluated to investigate the characteristics of torque or position transmission, based on either tube expansion or fluid compression.
However, changes in stiffness during operation were not considered.
When the stiffness of the HST was relatively high and the actuator load was low, the input and output at the tube tip remained synchronized. In such cases, the primary concerns involved response and backdrivability in leader--follower systems.
Spring stiffness, a parameter in the equation of motion, is generally treated as a constant determined by certain conditions, such as precharge pressure.
For example, Ganesh \textit{et al.}~\cite{ganesh2004dynamics} and Simonelli \textit{et al.}~\cite{simonelli2019hydrostatic} considered systems in which the leader and follower were double-acting cylinders connected by two fluid lines.
The fluid in the tube was approximated as a point mass, and the HST system was approximated as a two-degree-of-freedom mass-spring-damper model.
In another study, leader and follower cylinders were connected using a single tube~\cite{dong2019high}.
The two pistons of the cylinders and the fluid in the tube were then approximated as mass points, and a three-degree-of-freedom mass-spring-damper model was created to evaluate the HST system.

In other studies, remote actuators driven by HSTs have been used in human-worn exoskeletons, supernumerary limbs, and amusement robots.
However, tip state estimation was not considered in these applications~\cite{whitney2014low,veronneau2018high}.
Instead, efforts focused on increasing HST rigidity, under the assumption that the leader and follower moved synchronously.

Similar to this study, a water hydraulic actuator connected through a long tube for operation in harsh environments was investigated \cite{watanabe2024fault}. However, state estimation did not account for fluctuations in the stiffness of the hydraulic piping system.

Park \textit{et al.}~\cite{park2023stiffness} proposed a mechanism for actively switching the stiffness of a system.
The dynamics of a robot arm and the torque generated during contact with the external environment were evaluated under different stiffness conditions. However, no consideration was given to fluctuations in system stiffness caused by actuator load or to follower-side state estimation that accounts for such fluctuations. As the piping was short, the leader and follower moved almost synchronously, even when stiffness was set to ``low''.

{
\begin{table*}[htbp]
\centering
\setlength{\tabcolsep}{4pt}
\caption{Comparison with Previous Studies}
\label{table_comparison_study}
\begin{tabular}{llSSlc}
\toprule
\multirow{2}{*}{\textbf{Study}} &
\multirow{2}{*}{\textbf{\begin{tabular}{c} Applications  \\using HST \end{tabular}}} &
\multicolumn{2}{c}{\textbf{Tube dimensions}} & 
\multirow{2}{*}{\textbf{\begin{tabular}{c} Main study objective \end{tabular}}} &
\multirow{2}{*}{\textbf{\begin{tabular}{l} Fluctuating  \\ \ \ stiffness conditions \end{tabular}}}\\

& & {\begin{tabular}{c} Diameter \\ {[mm]} \end{tabular}} & {\begin{tabular}{c} Length \\ {[m]} \end{tabular}} \\
\midrule
Ganesh \textit{et al.}, 2004 \cite{ganesh2004dynamics}        &
MRI compatible system                    & 5 & 14 & Effects of tube dimensions on dynamics & \xmark\\

Whitney \textit{et al.}, 2014 \cite{whitney2014low}           &
Human interacting robot                        & 4.3 & 1.2 & Achieving full backdrivability & \xmark\\

V{\'e}ronneau \textit{et al.}, 2018 \cite{veronneau2018high}   &
Assistive exoskeleton robot                     & 2.3 & 8 & Responsiveness and transparency & \xmark\\

Guo \textit{et al.}, 2018 \cite{guo2018compact}                &
MRI compatible robot                     & 4 & 8 & Bidirectional and high rigidity & \xmark\\

Simonelli \textit{et al.}, 2019 \cite{simonelli2019hydrostatic}   &
Platform for MRI research                                & 4.3 & 10 & Multi-master intervention & \xmark\\

Dong \textit{et al.}, 2019 \cite{dong2019high}                  &
MR$^{\mathrm{(1)}}$ compatible motor                  & 4 & 10 & Reduced friction, high output & \xmark\\

Hyon and Akama, 2021 \cite{hyon2021air}                   &
Submersible robot                       & 3.2 & 20 & Torque control, robust & \xmark\\

Park \textit{et al.}, 2023 \cite{park2023stiffness}                   &
Human interacting robot                       & 6 & 1.1 & Stiffness-switching & \ \ \ \ \xmark$^{\mathrm{(2)}}$\\

This Study                                            &
Harsh-environment robot                  & 2.5 & 50 & Tip state estimation and control& \cmark\\

\bottomrule
\end{tabular}

\begin{flushleft}
\footnotesize{
(1) Magnetic resonance\\
(2) Switching settings}
\end{flushleft}
\end{table*}
}

\section{Method}
\subsection{Position estimation}
Several issues remain when applying sensorless hydraulic transmission to a long and flexible tube. The effective stiffness of the hydraulic transmission path varies with pressure and actuator load; residual or entrained air must be considered even after sufficient air bleeding; low-speed and low-impedance operation requires accurate regulation and measurement of small flow rates; actuator-side sensors and electronics cannot be assumed in harsh environments, therefore, estimation must rely on source-side pressure and displacement; and conventional HST synchronization assumptions fail because volumetric loss in a long and flexible tube is non-negligible.
Fig.~\ref{Power_and_information}(c) shows the system targeted in this study.
The actuator position was estimated under a quasi-static actuator-load balance.
As robots operating in high-risk environments move slowly and quasi-statically to maintain safety, this study employed a static model to estimate the actuator position.
Even when the subject is a real system containing dynamics, a static model is often effective for control.
As shown in Fig.~\ref{Base idea}, pressure within the transmission increases in response to actuator movement and/or load.
This increase in pressure caused the long tube to expand and the fluid inside the tube to compress, resulting in volumetric loss during transmission.
The volumetric displacement of the actuator is obtained by subtracting the volumetric loss during transmission from the volumetric displacement of the DFMC.
Therefore, the estimated actuator displacement $\hat{x}_2$ is calculated as,
\begin{align}
    \hat{x}_2=\frac{A_1}{A_2}x_1-\frac{\Delta V_\mathrm{hst}(p)}{A_2},  \label{eq_x2_hat}
\end{align}
where $x_1$ and $x_2$ represent the displacements of the DFMC and the actuator, respectively.
$A_{1}$ and $A_{2}$ represent the piston areas of the DFMC and actuator, respectively.
$\Delta V_\mathrm{hst}(p)$ represents the volumetric loss along the transmission path.
In the following sections, the gauge pressure is discussed.
The internal pressure $p$ is measured at the outlet of the DFMC.

\subsection{Model of volumetric loss during transmission} \label{Model_of_volume_changing}
Consider the change in the volume of the transmission path when the pressure inside the tube changes from the atmospheric pressure $p_\mathrm{o}$ to $p_\mathrm{i}$.
The change in volume of the transmission path is equal to the volumetric loss during transmission.
However, the volume change owing to negative pressure was neglected.
Assuming that the volume change in the transmission path is caused by the expansion of the tube and compression of the fluid within the tube, the change is calculated as,
\begin{align}
    \Delta V_\mathrm{hst} = \Delta V_\mathrm{t} + \Delta V_\mathrm{f}, \label{eq_Delta_V}
\end{align}
where $\Delta V_\mathrm{t}$ and $\Delta V_\mathrm{f}$ represent the volume changes owing to the deformation of the tube and the compression of the fluid, respectively.

First, tube deformation was considered.
When the volume of the tube at atmospheric pressure before deformation is $V_\mathrm{to}$ and the volume of the tube after deformation is $V_\mathrm{t}$, the volume change is, 
\begin{align}
    \Delta V_\mathrm{t} = V_\mathrm{t} - V_\mathrm{to}. \label{eq_Delta_V_t}
\end{align}
The inner radius of the tube before deformation is $r_\mathrm{i}$, and its length is $l$.
If the respective changes are $\Delta r_\mathrm{i}$ and $\Delta l$, then the volumes before and after deformation are, 
\begin{equation}
\begin{aligned}
        V_\mathrm{to} &= \pi r_\mathrm{i}^2l,  \\
        V_\mathrm{t}    &= \pi(r_\mathrm{i}+ \Delta r_\mathrm{i})^2(l+\Delta l)  \label{eq_V_t}. \\
\end{aligned}
\end{equation}
Additionally, assuming that the tube is isotropic and undergoes slight deformation in the elastic region, the stress--strain relationship is given as ~\cite{boresi2002advanced},
\begin{equation}
\begin{aligned}
\Delta r_\mathrm{i} &= \frac{r_\mathrm{i}}{E}\{\sigma_{\theta}-\nu(\sigma_{r} + \sigma_{z})\},\\
\Delta l            &= \frac{l}{E}\{\sigma_{z} - \nu(\sigma_{r} + \sigma_{\theta})\},
\end{aligned}
\label{eq_Delta_ri_l}
\end{equation}
where $E$ represents the Young's modulus of the tube, and $\nu$ represents the Poisson's ratio.
$\sigma_\mathrm{r},~\sigma_\mathrm{z},~\sigma_\mathrm{\theta}$ represent the radial, axial, and circumferential stresses on the inner radial surface of the tube, respectively.
Furthermore, when the pressure inside the tube is $pi$, the radial, circumferential, and axial stresses on the inner surface of the tube are given by the following thick-walled pipe equations~\cite{budynas1999advanced}:

\begin{equation}
\left\{
\begin{aligned}
        \sigma_\mathrm{r} &= -p_\mathrm{i},  \\
         \sigma_\mathrm{\theta} &= \frac{p_\mathrm{i}r_\mathrm{i}^2-p_\mathrm{o} r_\mathrm{o}^2-r_\mathrm{o}^2(p_\mathrm{o}-p_\mathrm{i})}{r_\mathrm{o}^2-r_\mathrm{i}^2},  \\
        \sigma_\mathrm{z} &= \frac{p_\mathrm{i}r_\mathrm{i}^2-p_\mathrm{o}r_\mathrm{o}^2}{r_\mathrm{o}^2-r_\mathrm{i}^2}  ,
\end{aligned}
\right.
\label{eq_sigma}
\end{equation}
where $r_\mathrm{o}$ is the initial outer radius of the tube and $p_\mathrm{o}$ is the pressure outside the tube.
From \eqref{eq_Delta_V_t}--\eqref{eq_sigma}, $\Delta V_\mathrm{t}$ can be calculated.

Subsequently, the compressive deformation of a fluid is considered.
When the volume of the fluid in the system at atmospheric pressure is $V_\mathrm{fo}$ and the volume after compression is $V_\mathrm{f}$, the volume change $\Delta V_\mathrm{f}$ is, 
\begin{align}
    \Delta V_\mathrm{f} = V_\mathrm{fo} - V_\mathrm{f}  .\label{eq_Delta_V_f}
\end{align}

Here, the fluid was modeled as water containing a trace amount of entrained air.
If the mixture volume ratio of air to water at atmospheric pressure is $\alpha$, the bulk modulus of elasticity of the fluid $\beta_\mathrm{f}$ is expressed as ~\cite{Matthiesen2018},
\begin{align}
    \beta_\mathrm{f} = \beta_\mathrm{w}\frac{1+\left(\frac{p_\mathrm{oa}}{p_\mathrm{ia}}\right)^{\frac{1}{\gamma}} \alpha}{1+\frac{\beta_\mathrm{w}}{\gamma p_\mathrm{ia}} \left(\frac{p_\mathrm{oa}}{p_\mathrm{ia}}\right)^\frac{1}{\gamma} \alpha},  \label{eq_beta_f} 
\end{align}
where $\beta_\mathrm{w}$ represents the bulk modulus of the liquid, $\gamma$ is the specific heat ratio, $p_\mathrm{ia} = p_\mathrm{i} + p_\mathrm{atm}$ and $p_\mathrm{oa} = p_\mathrm{o} + p_\mathrm{atm}$, and $p_\mathrm{atm}$ represents the atmospheric pressure.
Additionally, the bulk modulus of water was assumed to remain constant with pressure.
According to the definition of the bulk modulus, the relationship between the volume and pressure can be expressed as ~\cite{jelali2002hydraulic},
\begin{align}
    \frac{dV}{V} = -\frac{1}{\beta_\mathrm{f}}dp. \label{eq_dv_V}
\end{align}

Considering that the internal pressure changes from $p_\mathrm{o}$ to $p_\mathrm{i}$ and the fluid volume changes from $V_\mathrm{fo}$ to $V_\mathrm{f}$, integrating both sides of \eqref{eq_dv_V} yields, 
\begin{align}
    \int_{V_\mathrm{fo}}^{V_\mathrm{f}}\frac{dV}{V} &= -\int_{p_\mathrm{oa}}^{p_\mathrm{ia}}\frac{1}{\beta_\mathrm{f}}dp   \\
    V_\mathrm{f} &= V_\mathrm{fo} \mathrm{exp}\left(-\int_{p_\mathrm{oa}}^{p_\mathrm{ia}}\frac{1}{\beta_\mathrm{f}}dp\right)  \\  \label{eq_V_f}
\end{align}

$\Delta V_\mathrm{f}$ can be calculated using \eqref{eq_Delta_V_f}, \eqref{eq_beta_f}, and \eqref{eq_V_f}, where $V_\mathrm{fo}$ is calculated in advance as the sum of the tube, cylinder, and fitting volumes at atmospheric pressure.

\subsection{Identification and modification of model parameters}\label{identification_of_model_parameters}

The model parameters, namely, the gas-to-liquid volume ratio $\alpha$, Young's modulus of the tube $\mathrm{E}$, and Poisson's ratio $\nu$, are often unknown in practice.
Therefore, a method for identifying and applying these parameters is proposed.
In the identification setup, the actuator is fixed while the DFMC slowly displaces water so that the pressure--volume relationship can be recorded to determine $\alpha$ and $\mathrm{E}$.

\subsubsection{Identification of model parameters} 

The model parameters were identified according to the compact procedure summarized in Table~\ref{tab:identification_procedure}. In this study, $\nu=0.4$ was used because its influence on volumetric loss is small within the physically relevant range of 0--0.5, and this value is consistent with reported PA11 ($0.42$) and PA12 ($0.394$) nylon values~\cite{boisot2008failure,lirias467000}.
\newcounter{identify_step}
\setcounter{identify_step}{0}
\begin{table}[t]
\caption{Procedure for identifying model parameters}
\label{tab:identification_procedure}
\centering
\footnotesize
\begin{tabular}{c p{0.78\linewidth}}
\hline
\textbf{Step} & \textbf{Operation} \\
\hline
\refstepcounter{identify_step}\label{identify_1}\arabic{identify_step} & Connect the DFMC and actuator with a single tube and sufficiently bleed air from the hydraulic line. \\
\refstepcounter{identify_step}\label{identify_2}\arabic{identify_step} & Fix the actuator-side piston rod so that the actuator does not extend; the piston may be positioned at the stroke end. \\
\refstepcounter{identify_step}\label{identify_3}\arabic{identify_step} & Apply fluid pressure to the DFMC input port and slowly move the DFMC piston toward the actuator. As the actuator side is constrained, this displacement increases the tube pressure. \\
\refstepcounter{identify_step}\label{identify_4}\arabic{identify_step} & Treat the DFMC displaced volume as the transmission-path volume change at pressure $p$, plot the pressure--volume relation, and calculate the displaced volume as $\mathrm{A_1}x_1$ from the zero initial DFMC position. \\
\refstepcounter{identify_step}\label{identify_5}\arabic{identify_step} & Fit the measured data to the model in Section~\ref{Model_of_volume_changing} using nonlinear optimization to identify $\alpha$ and $\mathrm{E}$, while setting $\nu$ based on material information or a separately identified value. \\
\refstepcounter{identify_step}\label{identify_6}\arabic{identify_step} & Use the identified $\alpha$, $\mathrm{E}$, and $\nu$ for state estimation and control in actual operation (Section~\ref{Position_control}). \\
\hline
\end{tabular}
\end{table}

\subsubsection{Modification for other tubes} 

The model can be modified to accommodate tubes of the same material but with different lengths and diameters.

In water hydraulic actuators, air is intentionally removed from tubes and cylinders during assembly. However, eliminating air is difficult because small air pockets remain around cylinder edges and tube-fitting corners. The above method quantifies this residual air, which mainly remains in cylinders and fittings after bleeding, leaving negligible air in tubes. Thus, the air volume in the transmission path is regarded as nearly constant regardless of tube length or diameter.
Assuming that an HST system is constructed using a tube having an internal cross-sectional area of $A_\mathrm{A}$ and length of $L_\mathrm{A}$ at atmospheric pressure, and the air ratio identified using steps \ref{identify_1})--\ref{identify_5}) in Table~\ref{tab:identification_procedure} is $\alpha_\mathrm{A}$, the following relationship holds:
\begin{align}
    \alpha_\mathrm{A} &= \frac{V_\mathrm{ao}}{V_\mathrm{cy}+A_\mathrm{A} L_\mathrm{A}-V_\mathrm{ao}}, \\
    V_\mathrm{ao}   &= \frac{\alpha_\mathrm{A}}{1+\alpha_\mathrm{A}}(V_\mathrm{cy}+A_\mathrm{A} L_\mathrm{A}),
\end{align}
where $V_\mathrm{ao}$ is the air volume and $V_\mathrm{{cy}}$ is the internal volume of the cylinder and joint.
As the amount of air in the system when a tube with cross-sectional area $A_\mathrm{B}$ and length $L_\mathrm{B}$ is connected is estimated to be the same as $V_\mathrm{ao}$, the air ratio $\alpha_B$ can be expressed as, 
\begin{align}
    \alpha_\mathrm{B} &= \frac{V_\mathrm{ao}}{V_\mathrm{cy}+A_\mathrm{B} L_\mathrm{B}-V_\mathrm{ao}}, \\
             &= \frac{\frac{\alpha_\mathrm{A}}{1+\alpha_\mathrm{A}}(V_\mathrm{cy}+A_\mathrm{A} L_\mathrm{A})}{V_\mathrm{cy}+A_\mathrm{B}L_\mathrm{B}-\frac{\alpha_\mathrm{A}}{1+\alpha_\mathrm{A}}(V_\mathrm{cy}+A_\mathrm{A} L_\mathrm{A})}, \\
             &= \frac{\alpha_\mathrm{A}}{\frac{V_\mathrm{cy}+A_\mathrm{B}L_\mathrm{B}}{V_\mathrm{cy}+A_\mathrm{A}L_\mathrm{A}}(1+\alpha_\mathrm{A})-\alpha_\mathrm{A}}.
\end{align}
Therefore, based on $\alpha$ identified using a single tube and steps \ref{identify_1})--\ref{identify_5}) in Table~\ref{tab:identification_procedure}, a model can be constructed when tubes of different lengths and diameters are connected.

\begin{figure}[tbp]
\centering
  \includegraphics[width=0.9\linewidth]{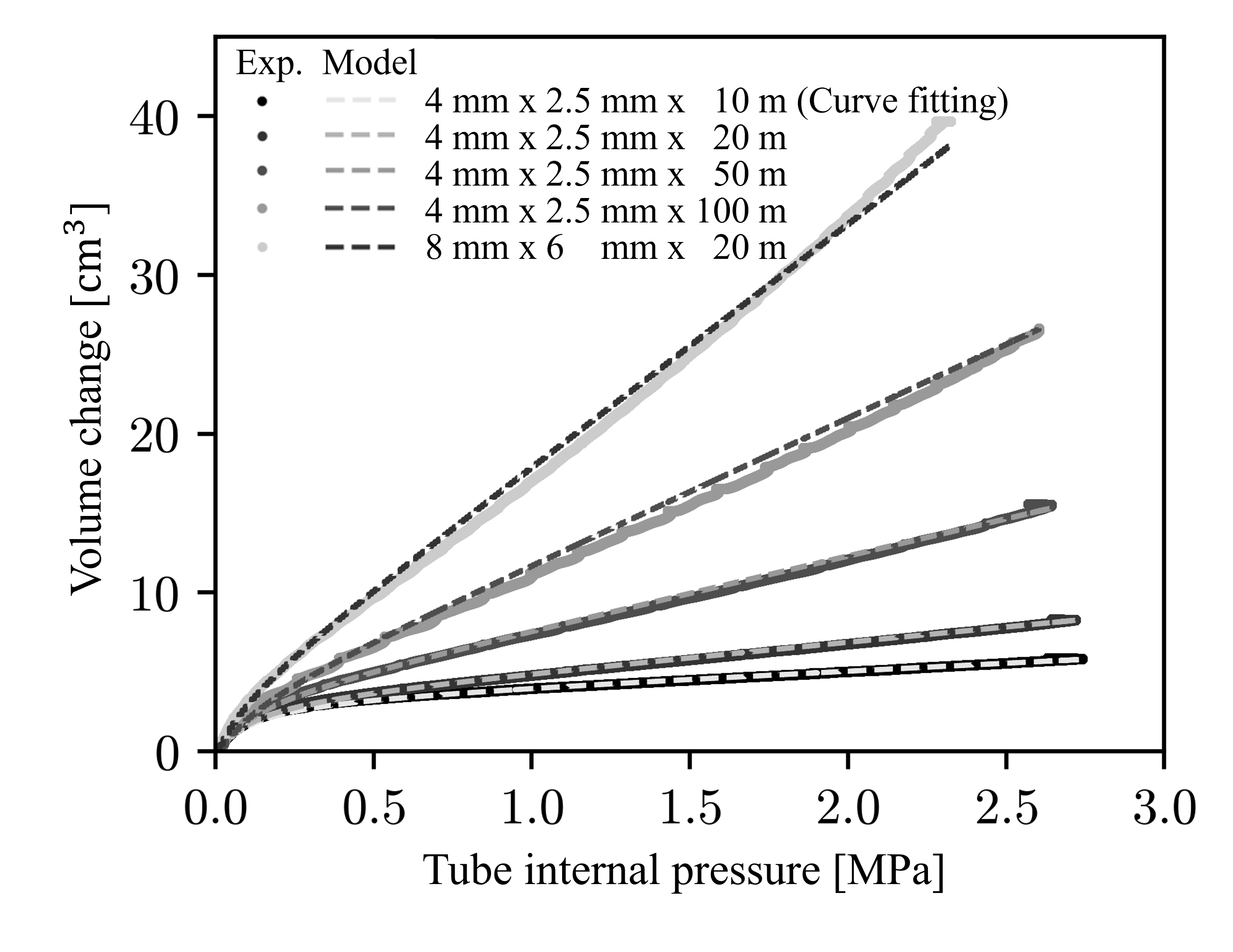}
  \caption{Volume change of tubes. Parameters were identified using ``4~mm $\times$ 2.5~mm $\times$ 10~m''. For other tubes, models were created based on those parameters.}
\label{comparison_tube}       
\end{figure}

Fig.~\ref{comparison_tube} shows a comparison of the modeled and actual measured values of the internal pressure and volume change for tubes of the same material but with different diameters and lengths, based on the parameters ($\alpha,E$) initially investigated.
For example, the legend ``4~mm $\times$ 2.5~mm $\times$ 10~m'' represents a tube with an outer diameter of 4 mm, an inner diameter of 2.5 mm, and a length of 10 m.
The initial parameter identification was performed for a tube with an inner diameter of 2.5 mm and a length of 10 m.
The results indicate that the calculated and measured volume changes with respect to $p$ generally agree for tubes of different lengths and diameters.

\begin{figure}[tbp]
\begin{minipage}[t]{0.30\linewidth}
    \subfloat{
        \overlaysubfig[height=2.5cm]{a}{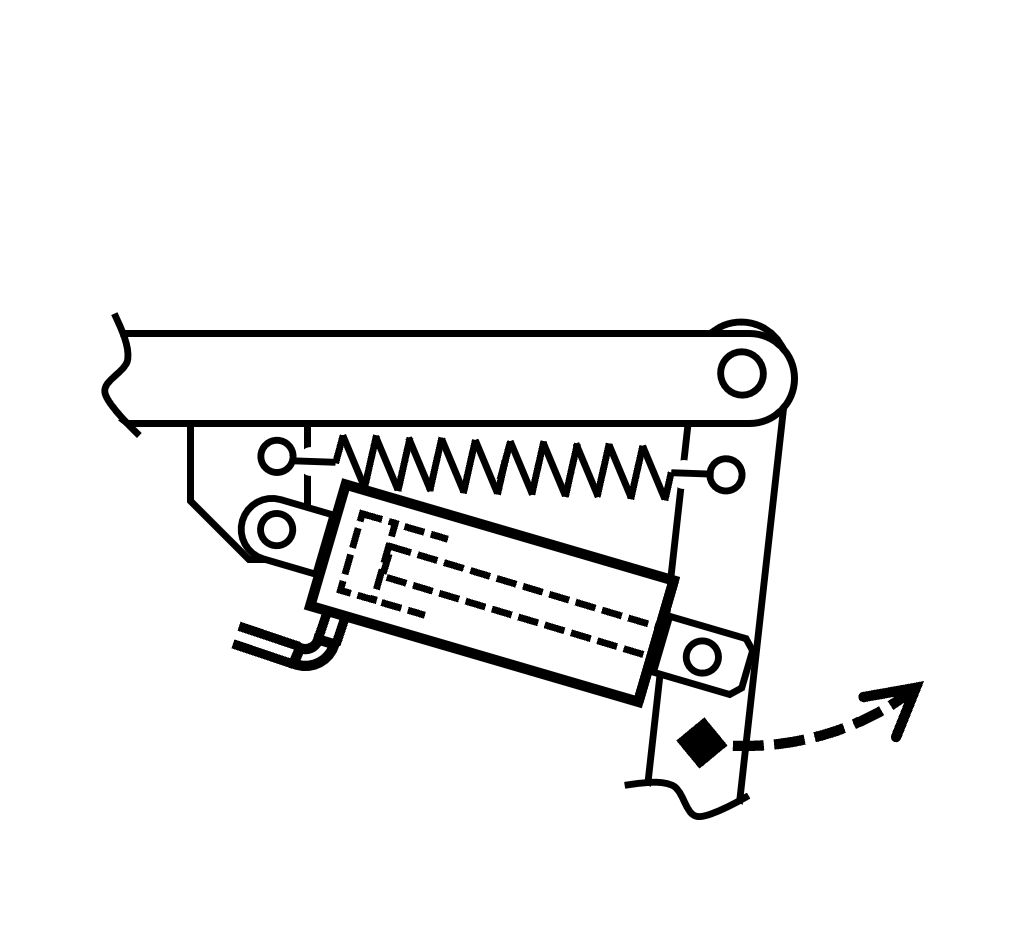}
        \label{Fig_robot_joint_a}
    }
\end{minipage}
\begin{minipage}[t]{0.30\linewidth}
    \subfloat{
        \overlaysubfig[height=2.5cm]{b}{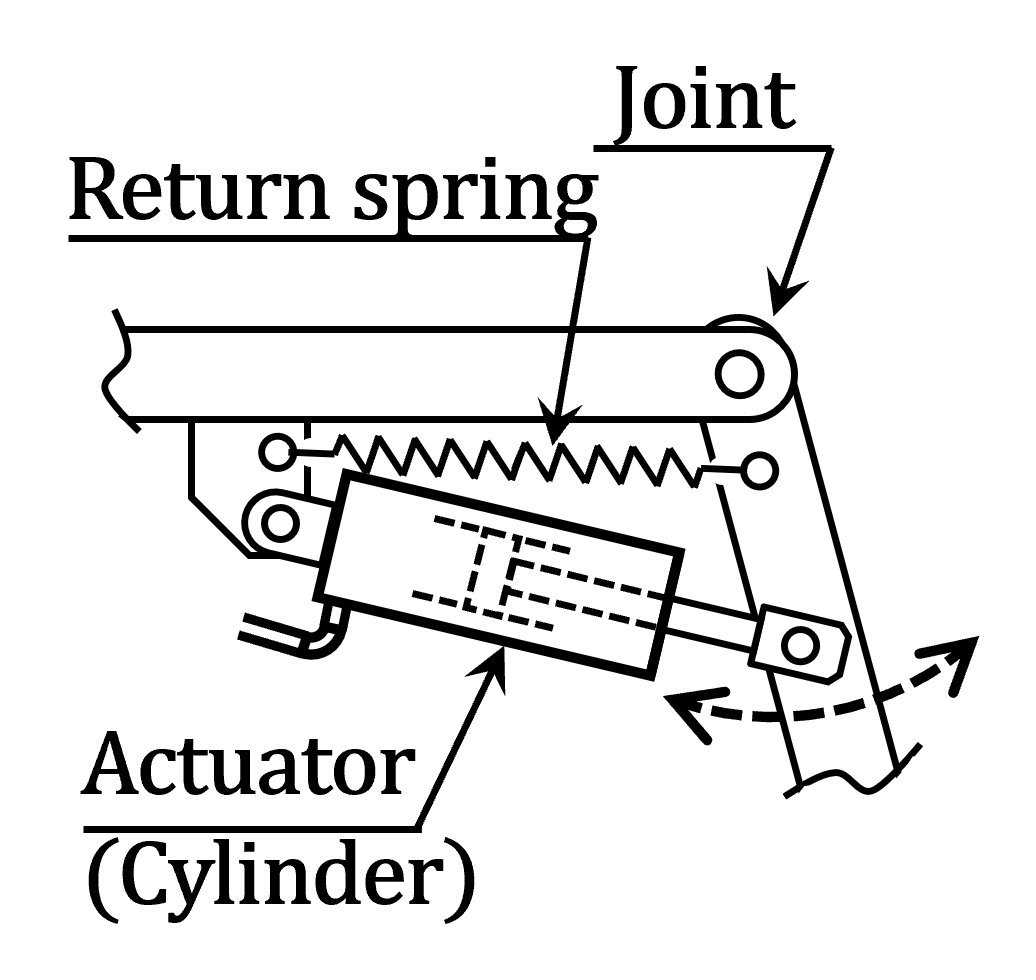}
        \label{Fig_robot_joint_b}
    }
\end{minipage}
\begin{minipage}[t]{0.32\linewidth}
    \subfloat{
        \overlaysubfig[height=2.5cm]{c}{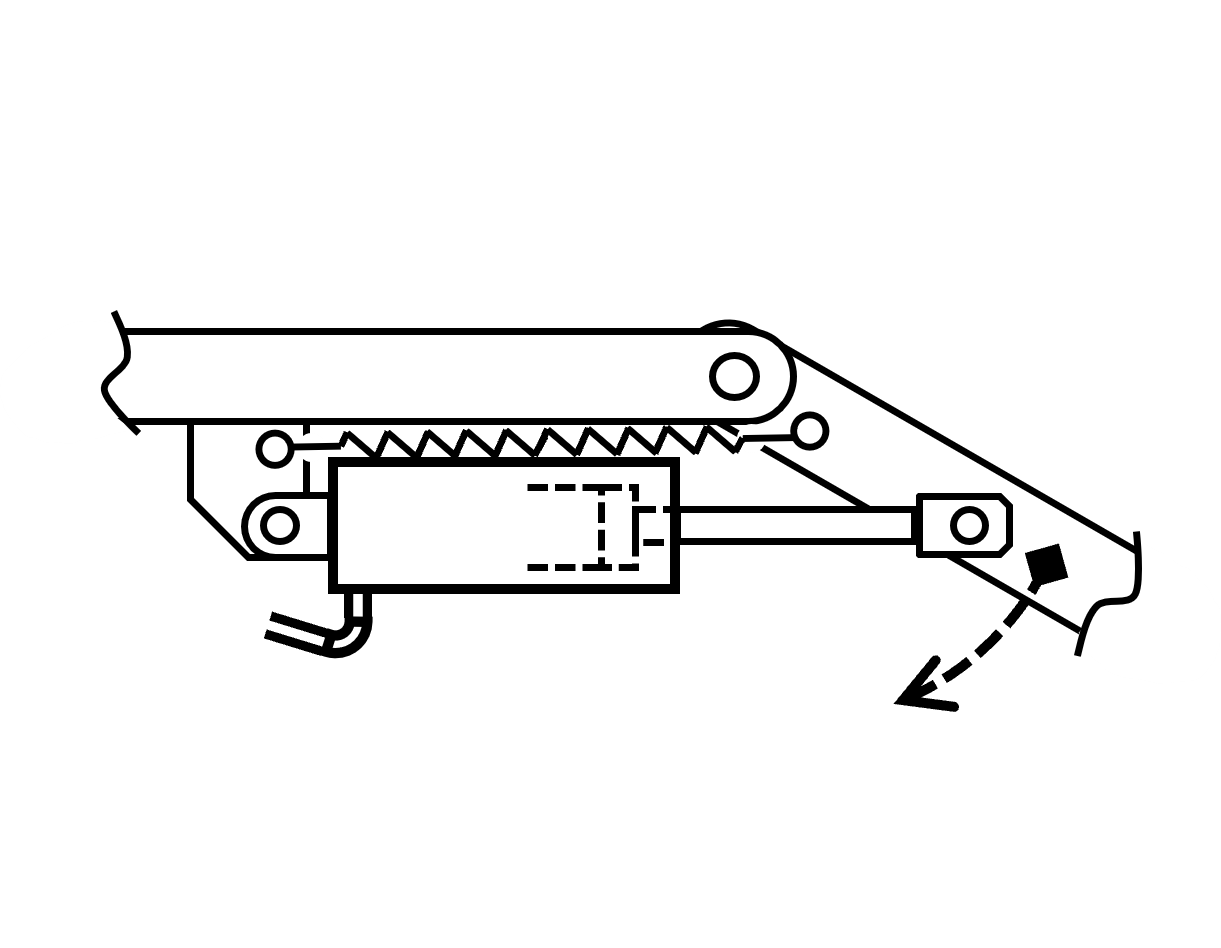}
        \label{Fig_robot_joint_c}
    }
\end{minipage}
\caption{Example joint configurations and mechanical end-stops. (a) Retracted EOS. (b) Mid-stroke. (c) Extended EOS. The re-identification was primarily validated through visual confirmation when the joint reached its mechanical limit. While hydraulic signatures can serve as a trigger, visual verification ensures the joint has reached its true limit rather than being obstructed by external obstacles.} 
\label{Fig_robot_joint}
\end{figure}

\begin{figure*}[!t]
\centering
\includegraphics[width=0.7\linewidth]{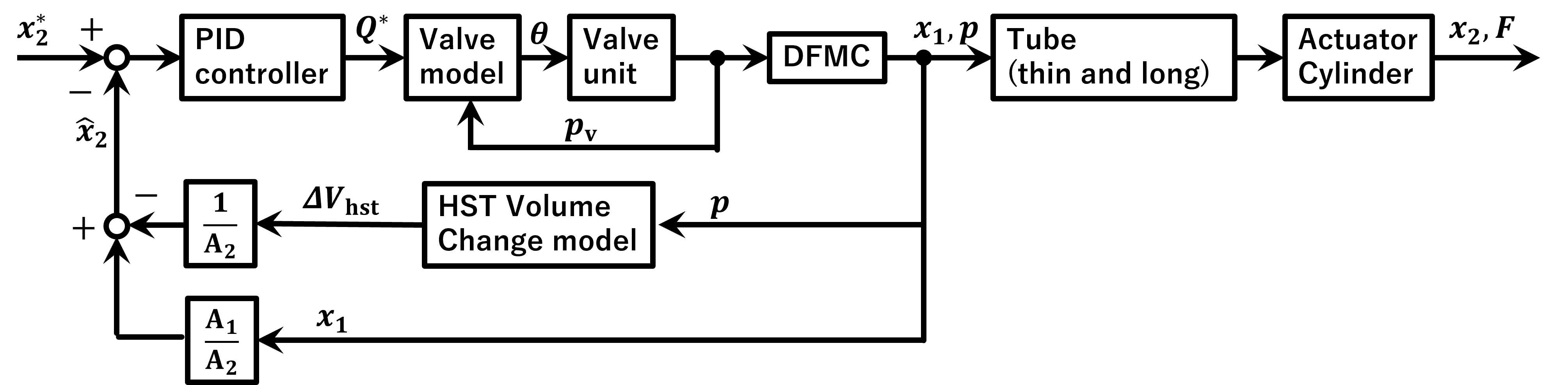}
\hfil
\caption{Block diagram of position control with compensation of the volume change.}
\label{block_diagram}
\end{figure*}

\subsection{In-situ parameter re-identification strategy}\label{In-situ reidentification}
The effective parameters E and $\alpha$ may drift due to environmental differences between the initial characterization site (e.g., a factory) and the actual deployment field, or gradual shifts during long-term operation (e.g., temperature changes due to weather). 
To help maintain estimation accuracy during tasks without requiring the robot to return from the hazardous environment, the SHAPE framework allows for in-situ parameter updating.

In the operational protocol, this process is performed during planned windows when visual information is reliable, such as before entering narrow spaces or during task transitions. 
By moving a joint to its mechanical end-of-stroke (EOS), the system replicates the fixed-actuator state used in the initial parameter identification.
While an abrupt pressure increase (hydraulic signature) can indicate contact, visual confirmation via cameras is the primary verification to ensure the joint has reached the true EOS rather than being obstructed by external obstacles.
Typical joint configurations and their corresponding EOS states are illustrated in Fig.~\ref{Fig_robot_joint}.

\section{System Implementation and Control}
\subsection{Valve unit and model} \label{valve_unit_and_valve_model}
\begin{figure}[tbp]
\begin{minipage}[t]{0.28\linewidth}
    \subfloat{
        \overlaysubfig[height= 3.5cm]{a}{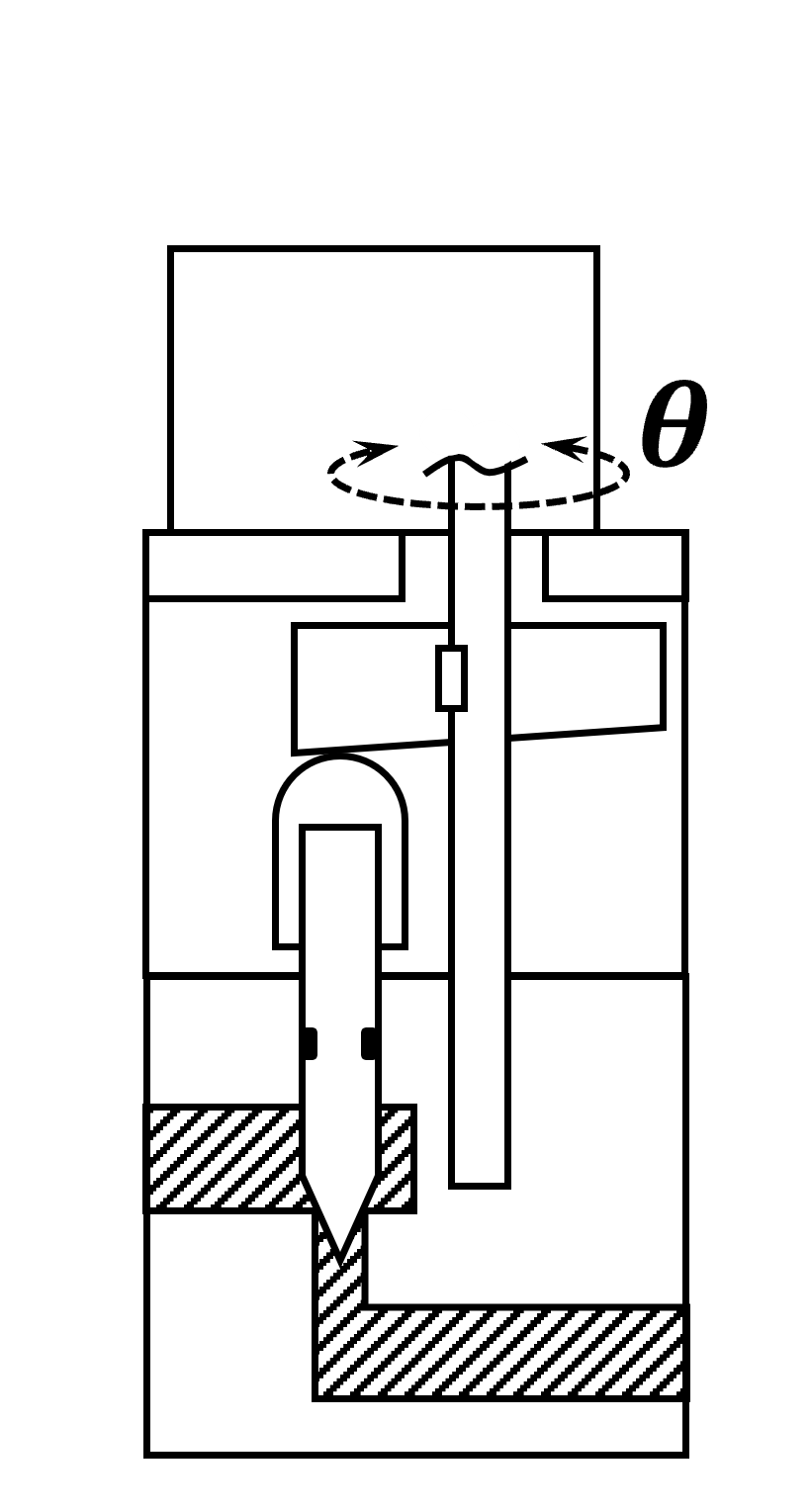}
        \label{Fig_valve_unit_a}
    }
\end{minipage}
\begin{minipage}[t]{0.36\linewidth}
    \subfloat{
        \overlaysubfig[height= 3.5cm]{b}{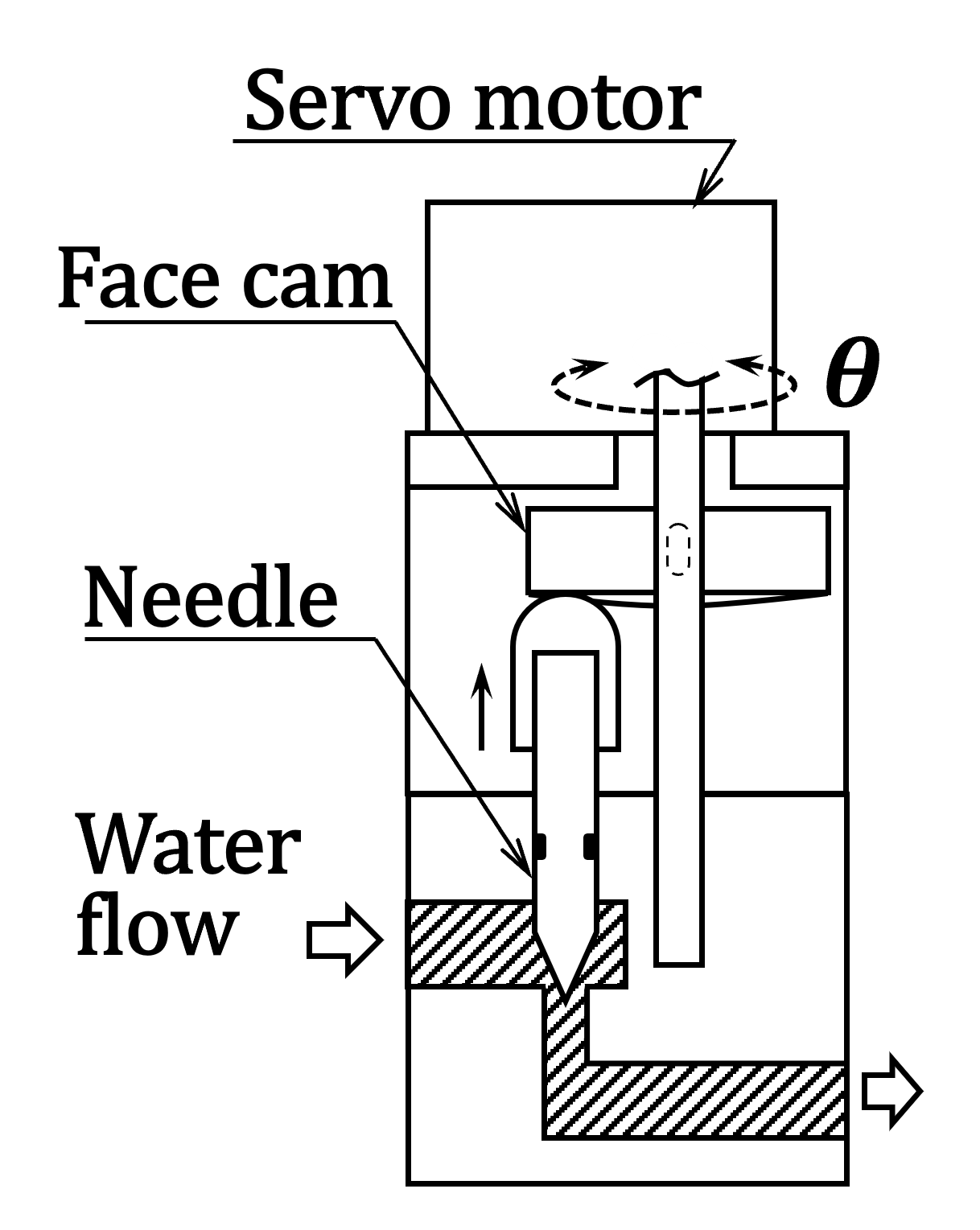}
        \label{Fig_valve_unit_b}
    }
\end{minipage}
\begin{minipage}[t]{0.32\linewidth}
    \subfloat{
        \overlaysubfig[height= 3.5cm]{c}{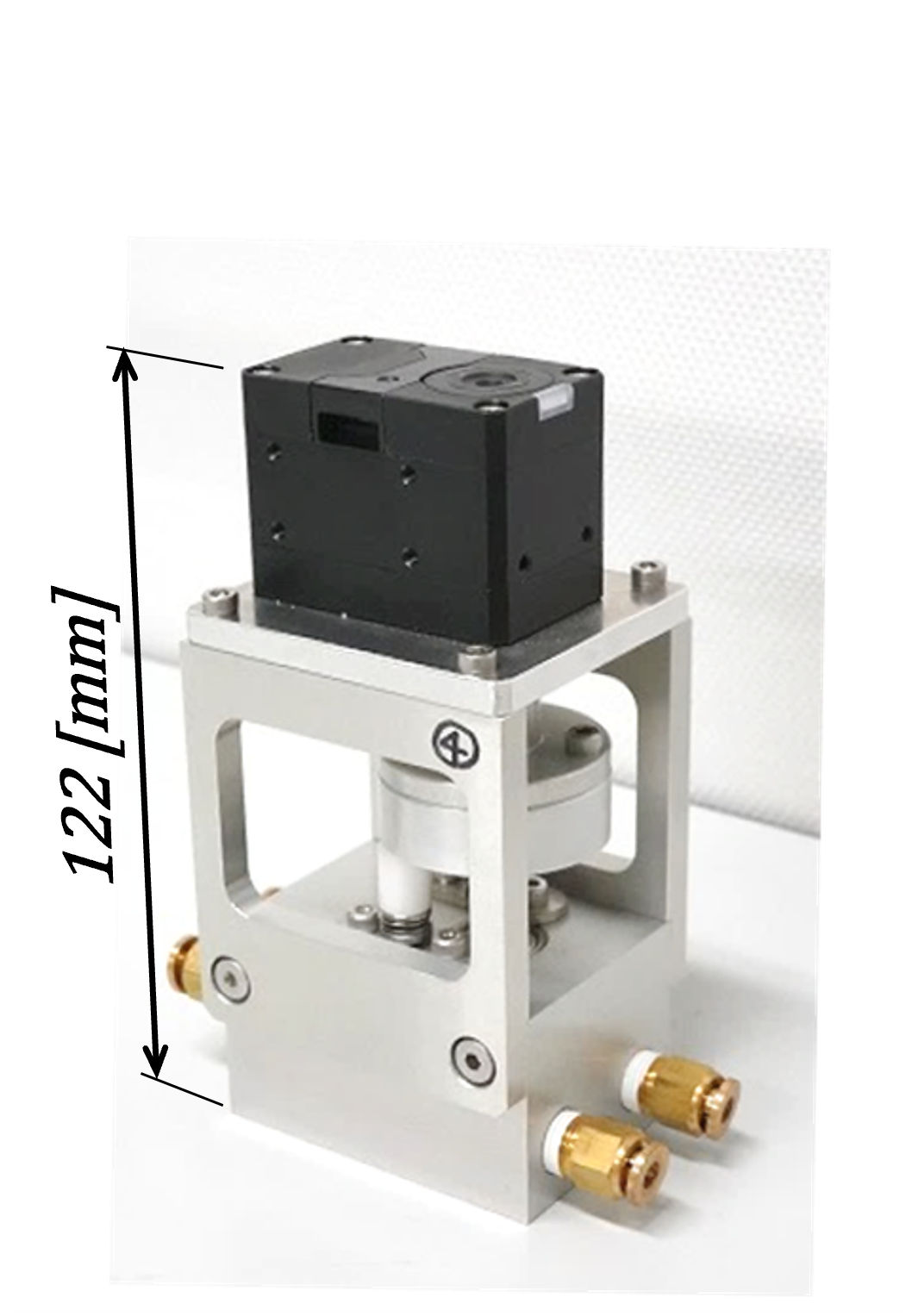}
        \label{Fig_valve_unit_c}
    }
\end{minipage}
\caption{Valve unit. (a) Schematic (valve closed). (b) Schematic (valve open). (c) Prototype appearance. The schematic shows only one valve. There are two valves in the depth direction which move in conjunction with the cam.}
\label{Fig_valve_unit}
\end{figure}

This study employed a custom-made flow-control valve unit with a Series Elastic Actuated Needle Valve (SEANV) structure~\cite{yoshimura2026series}. As illustrated in the schematic and prototype shown in Fig.~\ref{Fig_valve_unit}(a)--(c), the unit regulates water flow using a servo motor and a face cam to move two needles. While spool-type valves are conventional for water hydraulics~\cite{urata1998development}, reducing internal leakage to nearly zero is challenging due to the low kinematic viscosity of water, necessitating continuous operation even under static loads. By contrast, the proposed needle-type valve offers superior sealing performance and micro-flow control (Table~\ref{table_comparison_valve}), allowing the sensorless actuator to maintain intermediate positions effectively.
The relationship between the cam rotation angle $\theta$ and the state of the two valves ($\mathrm{wv_1,wv_2}$) is defined as, 
\begin{align*}
\theta_\mathrm{d2} \le \theta \le \theta_\mathrm{d1} &: \mathrm{wv_1} \text{ closed; } \mathrm{wv_2} \text{ closed.} \\
\theta_\mathrm{d1} < \theta &: \mathrm{wv_1} \text{ opens with } \theta; \ \mathrm{wv_2} \text{ closed.} \\
\theta < \theta_\mathrm{d2} &: \mathrm{wv_1} \text{ closed; } \mathrm{wv_2} \text{ opens as } \theta \text{ decreases.}
\end{align*}

To achieve the target flow rate $Q^\ast$ calculated by the controller, the command angle $\theta$ is determined using a pressure-dependent flow model~\cite{yoshimura2026series}:

\begin{align*} 
-Q_\mathrm{d} < Q^\ast < Q_d &: \theta = \theta_0 \in [\theta_\mathrm{d1}, \theta_\mathrm{d2}] \\ 
Q_d \leq Q^* &: \theta = \theta_\mathrm{d1}(p_\mathrm{p}, p_\mathrm{v}) + C_1 \left( \frac{|Q^\ast|}{\sqrt{p_\mathrm{p} - p_\mathrm{v}}} \right)^{\frac{1}{2}} \\
Q^\ast \leq -Q_d &: \theta = \theta_\mathrm{d2} (p_\mathrm{v}, p_\mathrm{atm}) + C_2 \left( \frac{|Q^\ast|}{\sqrt{p_\mathrm{v} - p_\mathrm{atm}}} \right)^{\frac{1}{2}}
\end{align*} 

where $Q_\mathrm{d}$ represents the dead zone, and $p_\mathrm{p},p_\mathrm{atm}$ and $p_\mathrm{v}$ denote the pump, tank (atmospheric), and system pressures, respectively.
The parameters $\theta_\mathrm{d_1},\theta_\mathrm{d_2}$ and $C_1,C_2$ are calibrated as functions of the pressures before and after each valve.

\begin{table}[tbp]
\centering
\caption{Comparison of Valve Types}
\label{table_comparison_valve}
\setlength{\tabcolsep}{6pt}
\begin{tabular}{p{50pt}ccc}
\hline
\textbf{Valve Type} & \textbf{Leak-proof} & \textbf{High Pressure} & \textbf{\begin{tabular}{c}Micro Flow\\Control\end{tabular}} \\ \hline
Spool  &   & + &   \\
Needle & + &   & + \\ \hline
\end{tabular}

\begin{flushleft}
\hspace{10pt}\footnotesize{+ : Advantageous}
\end{flushleft}
\end{table}

\subsection{Position feedback control} \label{Position_control}
The current tube-expansion model is formulated as a quasi-static pressure--volume model and does not explicitly include tube dynamics. Therefore, in this study, the estimated position $\hat{x}_2$ was used for closed-loop feedback control using a PID controller to validate whether the proposed estimation framework can support stable sensorless positioning. Fig.~\ref{block_diagram} shows the block diagram for actuator position control. The estimated position $\hat{x}_2$ is used as a feedback signal for a standard PID controller:
\begin{align} Q^\ast = K_\mathrm{p}e_\mathrm{ctrl} + K_\mathrm{i}\int e_\mathrm{ctrl} dt + K_\mathrm{d}\frac{de_\mathrm{ctrl}}{dt} \end{align}
where $e_\mathrm{ctrl} = x_2^\ast - \hat{x}_2$. This demonstrates that the SHAPE framework enables stable sensorless positioning under varying load conditions.

\section{Experiment}
The primary objective of the following experiments is not to demonstrate an advanced control algorithm, but to verify that the proposed SHAPE framework provides accurate position estimation sufficient to enable stable sensorless feedback control. To this end, a basic PID controller was employed as a representative application. The dynamic tube model discussed in Section~\ref{Discussion} provides a simple representation of temporal delay and transient pressure behavior. Incorporating time dynamics into control could enable future two-degree-of-freedom-like strategies that combine model-based feedforward with feedback correction. However, modeling tube dynamics with sufficient accuracy remains a nontrivial problem and is an important subject for future work. This choice demonstrates that the SHAPE framework provides a baseline for achieving stable final positioning even when using a standard control law.

\subsection{Experimental setup}
\begin{figure*}[tbp]
\centering

\subfloat{
    \overlaysubfig[width=0.85\linewidth]{a}{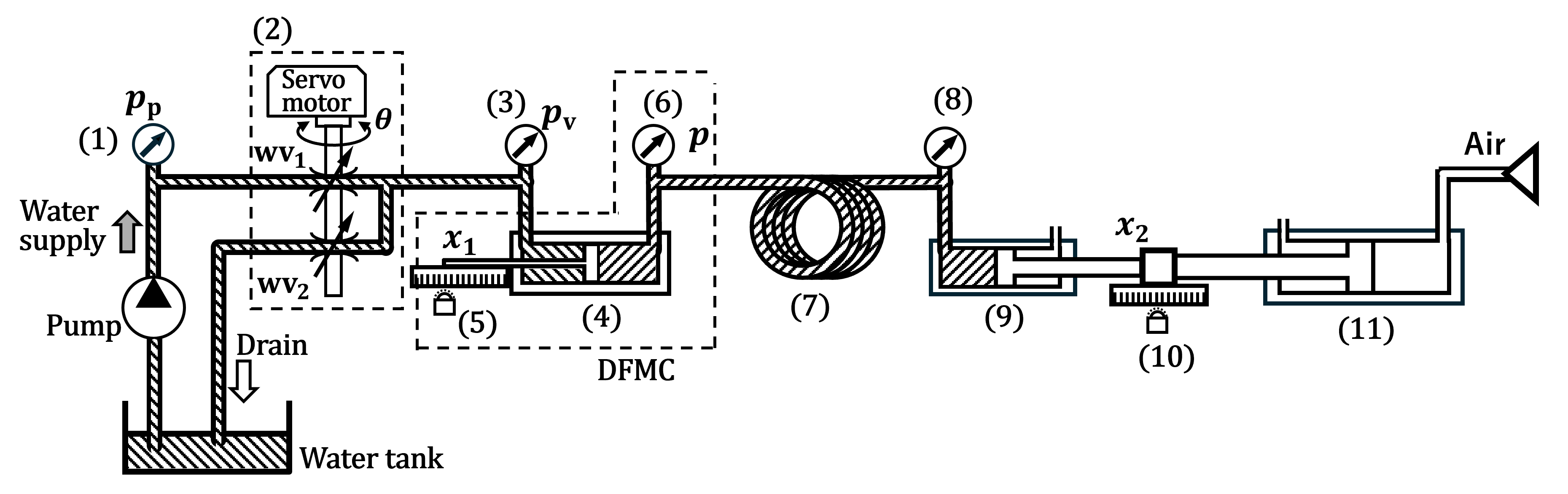}
    \label{Exp_eq_a}
}\\[1ex] 
\subfloat{
    
    \overlaysubfig[width=0.85\linewidth]{b}{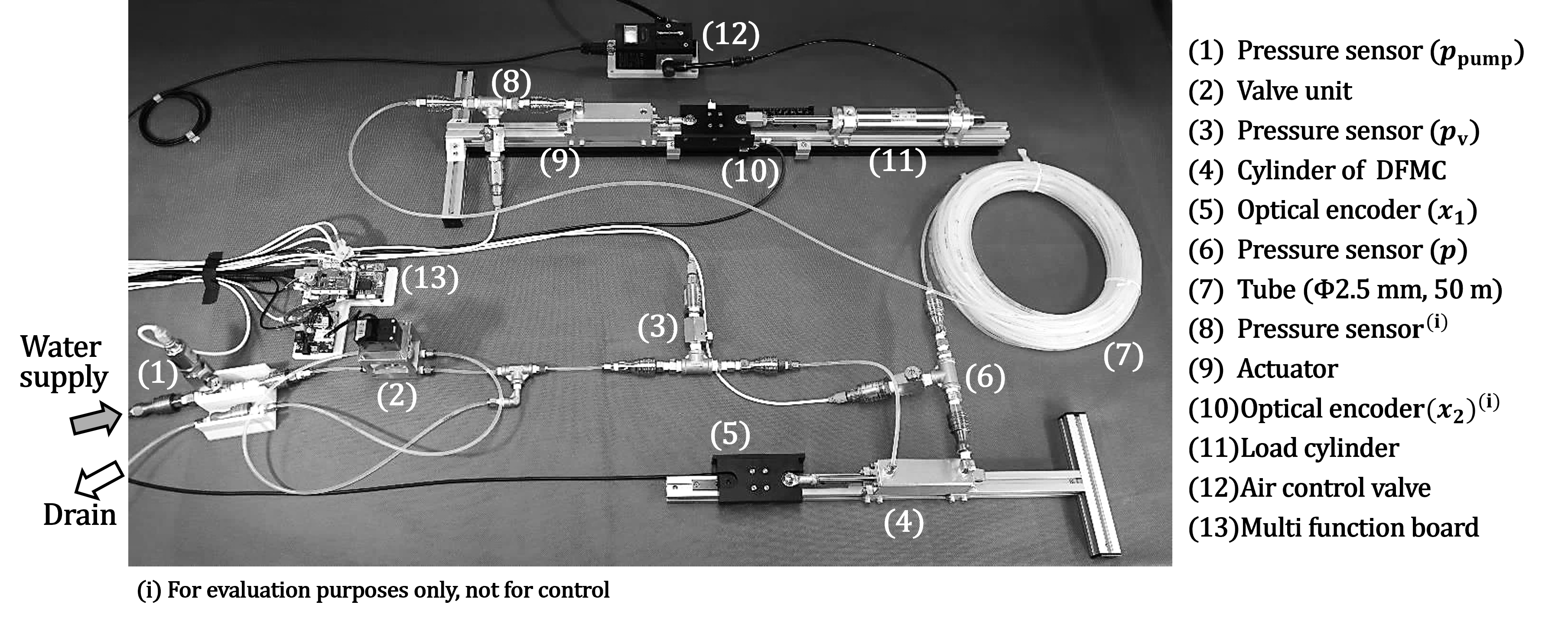}
    \label{Exp_eq_b}
}

\caption{Experimental setup of the proposed SHAPE system.
(a) Schematic showing valve unit, DFMC, 50~m transmission tube, and actuator with load cylinder (Pneumatic).
(b) Photograph of the experimental setup.}

\label{Experimental_equipment}
\end{figure*}

Figs.~\ref{Experimental_equipment}\subref{Exp_eq_a} and \ref{Experimental_equipment}\subref{Exp_eq_b} show a schematic and photograph of the experimental system, respectively. Water pressurized by an electric pump (maximum 3.4 MPa) was delivered to the valve unit described in Section \ref{valve_unit_and_valve_model}. The DFMC output port was connected to the controlled actuator via a 50~m flexible thin tube with an inner diameter of 2.5 mm. While the DFMC piston displacement and output pressure were measured for feedback control via an optical linear encoder and a pressure sensor, an additional encoder and pressure sensor were temporarily attached to the actuator solely for experimental evaluation. 

The actuator was a custom single-acting cylinder without a return spring to evaluate the most fundamental configuration; thus, its unpressurized motion depended entirely on the pneumatic cylinder load, which was applied opposite to the actuator output direction. Detailed equipment specifications and common variable values are summarized in Tables~\ref{table_equipment} and \ref{Variable_values}, respectively.

\begin{table}[tbp]

\centering
\caption{Control system equipment}
\label{table_equipment}
\begin{tabular}{lll}
\hline
 \textbf{Equipment} &  \textbf{Model Name} & \textbf{Manufacturer}    \\ \hline
Tube                & N2-4-4 × 2.5      &  Niita Corp.      \\ 
Electric pump       & KY-20A            &  Kyowa Inc.          \\ 
Valve unit          & ADSA-22121-01r1   &  Chugai technos      \\ 
Motor for valves & XM430-W350    &  Robotis              \\ 
Optical encoder     & AEDR-8710-102     &  Broadcom Inc.         \\ 
Pressure sensor     & PSE577-02         &  SMC Corp.          \\ 
Cylinder for DFMC   &piston:$\phi16$, rod:$\phi8$ & Chugai technos \\
Cylinder actuator   &piston:$\phi16$, rod:$\phi8$ & Chugai technos \\
Load actuator       & CM2C32-100Z-NV    &  SMC Corp.         
\\ \hline
\end{tabular}
\end{table}

\begin{table}[tbp]
\centering
\caption{System Parameters}
\label{Variable_values}
\begin{tabular}{lcSc} \toprule
\textbf{Fixed Parameters} & \textbf{Variable} & \textbf{Value} & \textbf{Unit} \\ \midrule
Tube inside radius & $r_\mathrm{i}$ & 1.25 & mm \\
Tube outside radius & $r_\mathrm{o}$ & 2.00 & mm \\
Tube Length & $l$ & 50 & m \\
Piston area  & $A_1, A_2$ & 201 & $\mathrm{mm^2}$ \\ 
Water viscosity & $\mu$  & 1.0e-3 & $\mathrm{Pa \cdot s}$ \\ \midrule
\multicolumn{4}{l}{\textbf{Identified Effective Parameters}} \\ \midrule
Tube Young's modulus & $E$ & 287 & MPa \\
Entrained air ratio & $\alpha$ & 0.0159 & -- \\
Poisson's ratio (fixed) & $\nu$ & 0.4 & -- \\ \bottomrule
\end{tabular}
\end{table}

\subsection{Position control experiments}\label{experimental_condition}
In the experiment on remote position control of a sensorless actuator using the proposed method, the target position was changed stepwise under several load conditions. 
The target position, in millimeters, was varied as follows: \\
$ 0 \rightarrow 30 \rightarrow 15 \rightarrow 30 \rightarrow 33 \rightarrow 30 \rightarrow 0$.\\
The load conditions were as follows:
\begin{enumerate}
    \item[(a)] A load of 241 N was applied continuously.
    \item[(b)] A load of 321 N was applied continuously.
    \item[(c)] The load was alternated between 241 and 321 N \\every 30 s.
\end{enumerate}

The common settings were as follows:
The frequency of state estimation was set to 100 Hz, and the frequency of the angle commands to the servo motor for the valve unit was set to 10 Hz.
The servo motor was driven according to the command value via the built-in driver of the XM430-W350 (Robotis).
For the PID controller shown in Fig.~\ref{block_diagram}, the proportional gain was set to $K_\mathrm{p}=100$, the integral gain was set to $K_\mathrm{i}=5$, and the differential gain was set to $K_\mathrm{d}=50$.
However, in the anti-windup scheme, the integral value of the deviation was limited to $\pm50~\mathrm{mm}$.
Additionally, a dead band was set in the range of $\pm1~\mathrm{mm}$. Within the deadband, the valve cam rotation angle was set to degrees at $\theta_0 = \theta_\mathrm{d1}-3$.
A Savitzky--Golay filter was applied to the data from the pressure sensors used for control.

\subsection{Results}

\begin{figure*}[tbp]

\begin{minipage}[t]{0.32\linewidth}
    \subfloat{
        \overlaysubfig[height=12.2cm]{a}{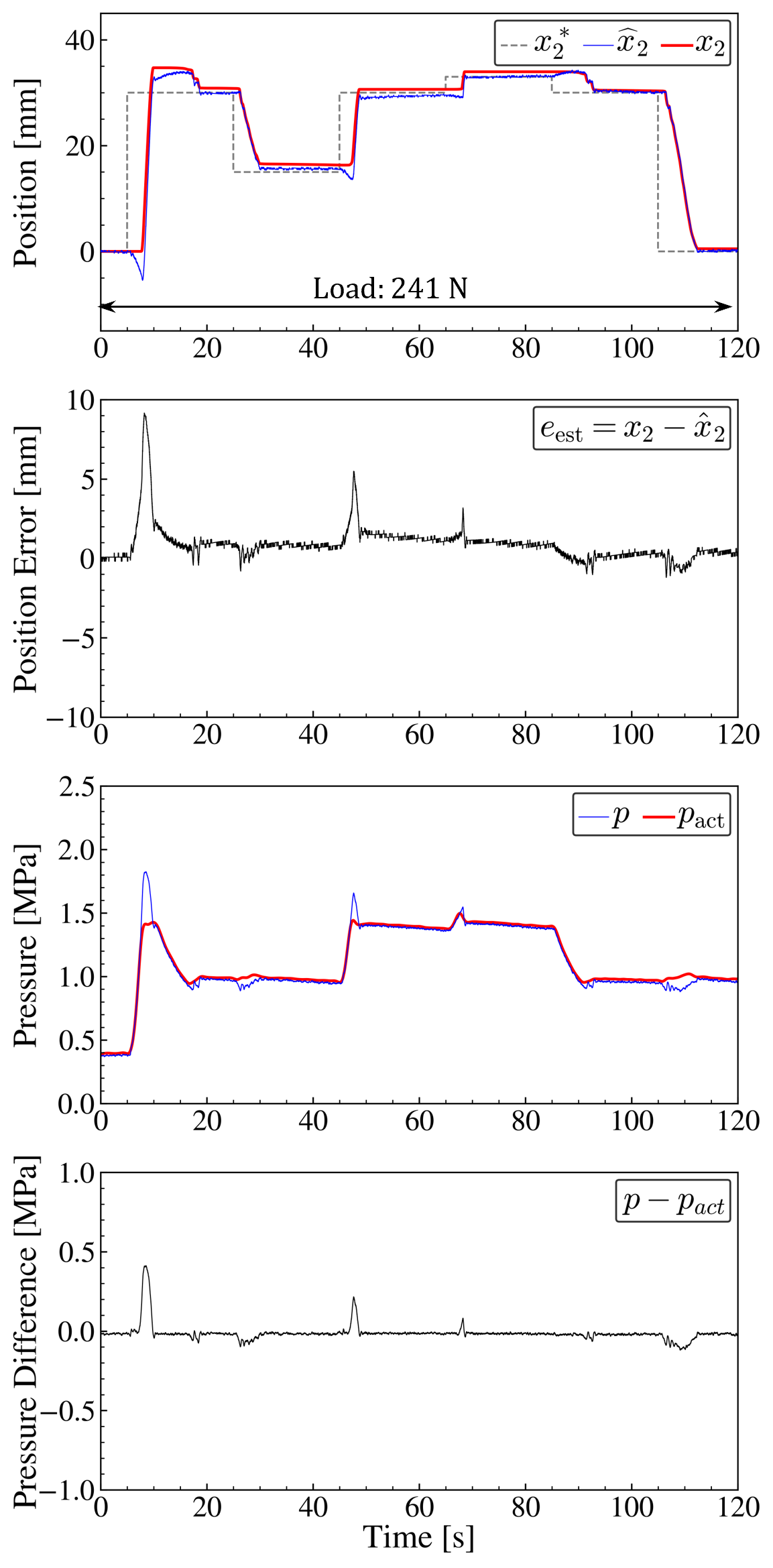}
        \label{result_a}
    }
\end{minipage}
\hspace{8pt}
\begin{minipage}[t]{0.32\linewidth}
    \subfloat{
        \overlaysubfig[height=12.2cm]{b}{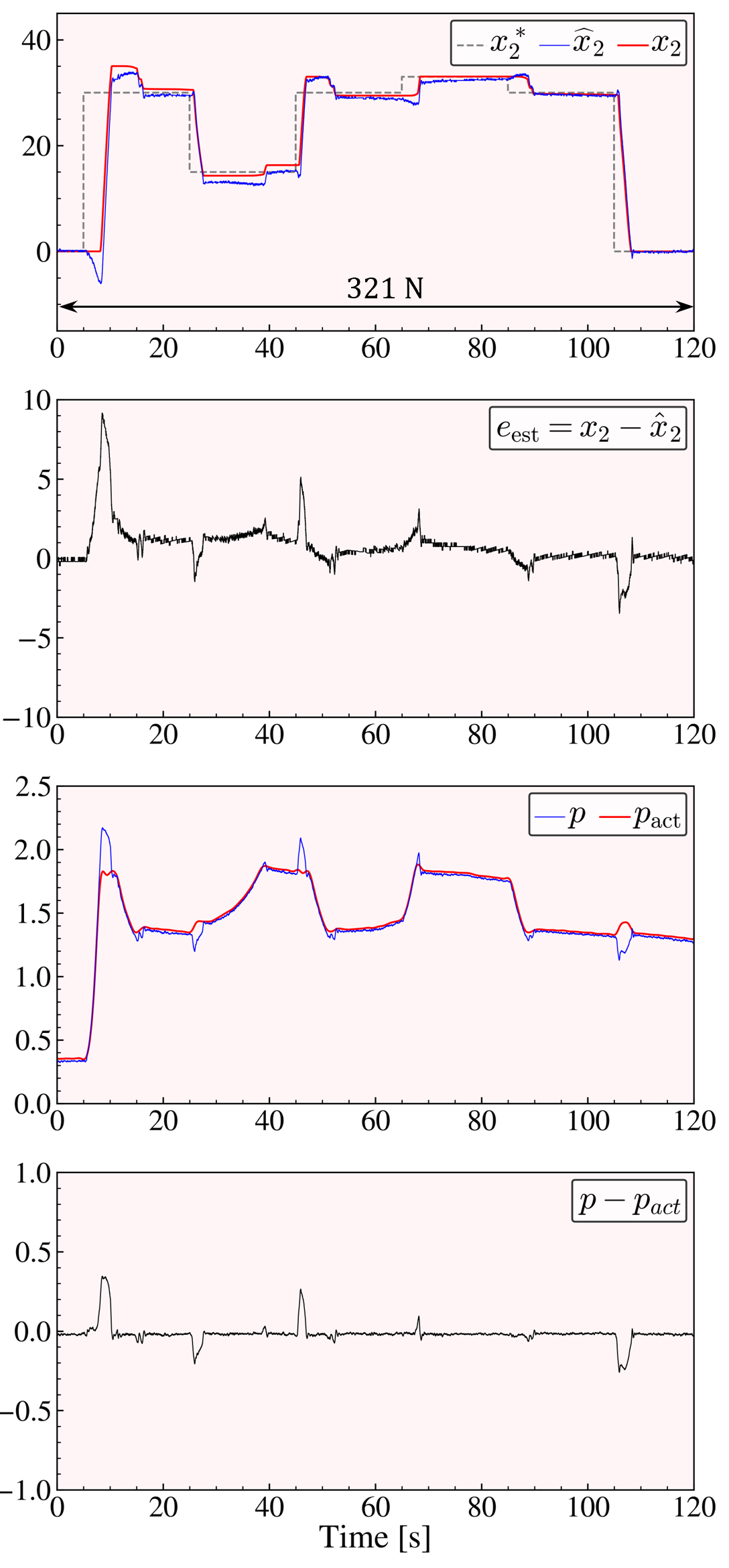}
        \label{result_b}
    }
\end{minipage}
\begin{minipage}[t]{0.32\linewidth}
    \subfloat{
        \overlaysubfig[height=12.2cm]{c}{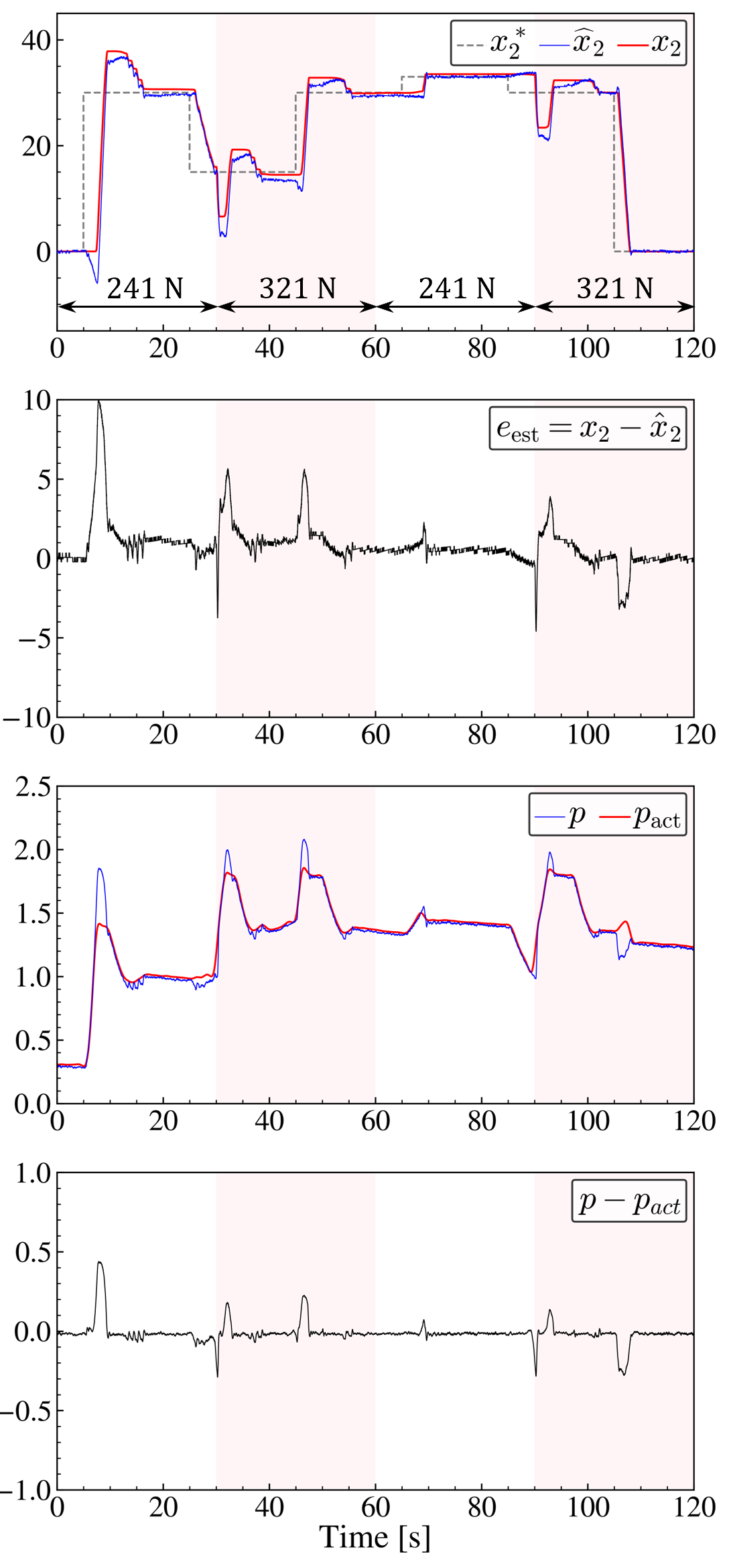}
        \label{result_c}
    }
\end{minipage}
\caption{Experimental results. Top row: position; second row: position error; third row: pressure; bottom row: pressure difference (all time-series). Pink regions indicate a 321 N load. (a) Continuous load of 241 N. (b) Continuous load of 321 N. (c) Load alternating between 241 and 321 N.}
\label{Experimental_results}
\end{figure*}

Fig.~\ref{Experimental_results} shows the time-series results of the positioning control experiments under load conditions (a), (b), and (c) in Section \ref{experimental_condition}, where each measurement plots sensor-acquired values filtered by a Butterworth low-pass filter.

Furthermore, Table~\ref{tab:estimation_errors} lists the Root Mean Square Error (RMSE) and Maximum Error for $e_\mathrm{est}$ and $e_\mathrm{base}$ computed over the entire experimental interval, including transient and load-switching periods. Here, $e_\mathrm{base}$ represents the baseline estimation error when volumetric loss in the transmission line is entirely neglected.

As shown in the top graphs of Fig.~\ref{Experimental_results}, the estimated value $\hat x_2$ exhibited a transient misestimation in the opposite direction upon actuator startup; however, the true actuator position $x_2$ closely followed the target $x_2^{\ast}$ throughout the trial.

Additionally, the bottom graphs indicate that although transient discrepancies occur, the proximal pressure $p$ and actuator-side pressure $p_{\mathrm{act}}$ promptly converge toward agreement whenever the velocity decreases toward a stop.

\begin{table}[t]
\centering
\footnotesize
\begin{threeparttable}
\caption{Estimation Errors under Each Condition [$\mathrm{mm}$]}
\label{tab:estimation_errors}
\begin{tabular}{llccc} 
\hline
\textbf{Item} & \textbf{Metric} & \textbf{Cond. (a)} & \textbf{Cond. (b)} & \textbf{Cond. (c)} \\ \hline
\multirow{2}{*}{$e_\mathrm{est}$\tnote{$\dagger$1} }  & RMSE & \phantom{0}1.414 & \phantom{0}1.495 & \phantom{0}1.664 \\
 & Max. error & \phantom{0}9.157 & \phantom{0}9.157 & 10.057 \\ \hline
\multirow{2}{*}{$e_\mathrm{base}$\tnote{$\dagger$2} }  & RMSE & 32.039 & 40.286 & 36.804 \\ 
 & Max. error & 40.752 & 50.944 & 49.861 \\ \hline

\end{tabular}
\begin{tablenotes}[flushleft] 
    \item[$\dagger$1] $e_{est} = x_2 - \hat{x}_2$
    \item[$\dagger$2] $e_\mathrm{base} = x_2 - (A_1/A_2)x_1$

\end{tablenotes}
\end{threeparttable}
\end{table}

\section{Discussion}
\label{Discussion}

The results in Fig.~\ref{Experimental_results} demonstrate that the SHAPE framework enables accurate position estimation and stable feedback control under varying load conditions. Although the estimation error $e_{est}$ increases at startup, the pressures at both ends ($p$ and $p_{act}$) quickly equalize during movement, driving the estimated position $\hat x_2$ toward the true value $x_2$.
Under the conventional HST assumption that neglects line volumetric losses, source-side displacement is directly equated to actuator displacement.
However, Table~\ref{tab:estimation_errors} shows that this assumption leads to massive discrepancies, with the maximum $e_\mathrm{base}$ reaching 40--51 mm. These errors were too large to plot alongside the compensated results, confirming that precise long-distance remote control is impractical without compensation.
By contrast, the SHAPE method compensates for real-time volumetric losses, enabling accurate target positioning without actuator-side sensors.

\subsection{Major contributions and practical significance}
The primary contribution of this study is the establishment of an enabling framework for remote and sensorless position control through thin and long flexible tubes.
By explicitly modeling pressure-dependent volumetric loss, the SHAPE framework overcomes the need for actuator-side electronics, which are highly susceptible to failure in hazardous environments, such as radiation-exposed or underwater areas.

A key technical advantage of the proposed system is its inherent low mechanical impedance.
While this characteristic may limit high-speed response, it is exceptionally beneficial for robots operating in uncertain hazardous environments.
The compliance of the long-tube hydraulic drive provides passive safety, allowing the robot to mechanically absorb shocks from unexpected contact or collisions.
In such field operations, ensuring safety through low-impedance interaction is often a higher priority than achieving high-precision force control.

SHAPE enables closed-loop position control within this compliant framework without requiring actuator-side sensors.
Furthermore, the in-situ parameter re-identification strategy provides an on-site procedure for refreshing model parameters ($\mathrm{E}$ and $\alpha$) without actuator-side sensors, using mechanical end-stops and source-side DFMC measurements.
In this study, its effectiveness is supported by sensorless position-control results obtained through closed-loop feedback control in the tested setup, rather than by a separate long-term validation dataset.

\subsection{Limitations and future challenges}
Despite its effectiveness as a baseline implementation, the current SHAPE framework has several limitations that define directions for future work:
\subsubsection{Transient dynamics} 
\begin{figure}[tbp]
\centering
  \includegraphics[width=0.95\linewidth]{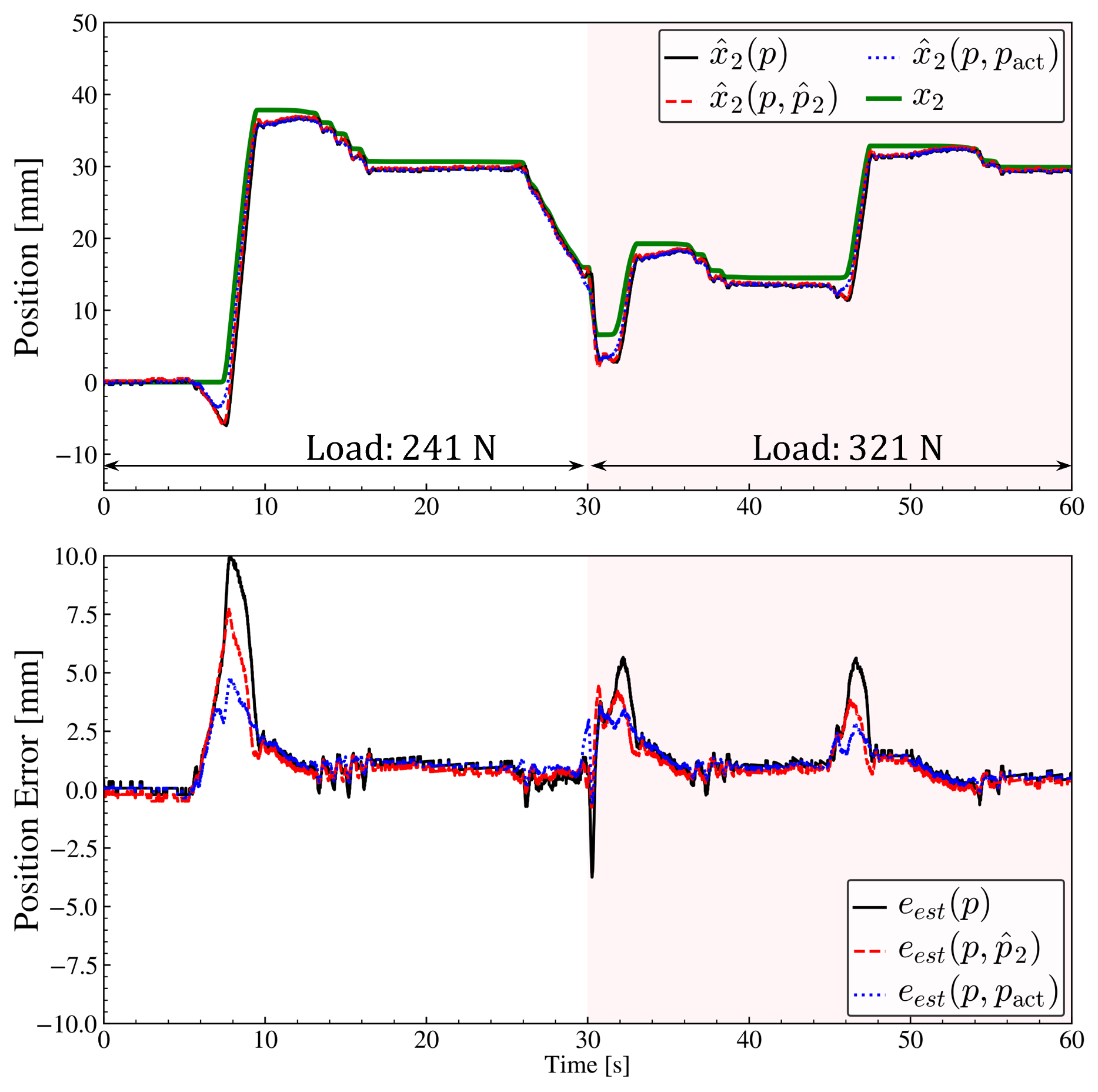}

  \caption{Comparison of position estimation with and without actuator-side pressure information.}
\label{comparison of estimation}       
\end{figure}

The present tube-expansion model is quasi-static, which was generally valid in Fig.~\ref{Experimental_results} because the measured proximal pressure \(p\) and actuator-side pressure \(p_{\mathrm{act}}\) were nearly identical in most data.
However, during sudden start/stop motions, viscous resistance and pressure-propagation delay can create a pressure gradient that the present pressure--volume model cannot represent.
Appendix~\ref{app:two_segment_pressure_propagation} describes a two-segment pressure-propagation model as a minimal extension for addressing this limitation.
It was investigated whether this simple actuator-side pressure estimate could improve position estimation.
Fig.~\ref{comparison of estimation} and Table~\ref{tab:comparison_error_from_pressure} compare position estimation using only proximal pressure with cases using either estimated pressure $\hat p_2$ or measured pressure $p_\mathrm{act}$ in addition to DFMC information ($x_1$ and $p$).
This simulation used load condition (c) described in Section~\ref{experimental_condition}.
Here, the pressure $\hat p_2$ was obtained using a Kalman filter with the two-segment model in Appendix~\ref{app:two_segment_pressure_propagation}. In the Kalman filter, $x_1$ and $p=p_1$ were the observed values;
$\hat x_2(p,\hat p_2)$ and $\hat x_2(p,p_\mathrm{act})$ are the estimated actuator positions obtained by dividing the long tube into two sections whose pressures are $p$ and either $\hat p_2$ or $p_\mathrm{act}$, respectively. Fig.~\ref{comparison of estimation} shows the first 60 s of load condition (c).

A limitation of this analysis is that the estimated actuator-side pressure was evaluated retrospectively using data obtained under the one-state control model and was not directly incorporated into the feedback control loop. As the model parameters were tuned with reference to the measured actuator-side pressure in retrospective analysis, validation under different tube lengths, load profiles, and operating conditions is required to evaluate the generalizability of actuator-side pressure estimation.

These results indicate that estimating actuator-side pressure improves position estimation, which can improve position control performance and provide a path toward secondary force-information estimation.
Further improvement may be achieved by increasing the number of tube segments to approximate a distributed pressure field, as studied for pneumatic pressure propagation~\cite{10547707}. The present extension captures pressure propagation through distributed compliance; however, pressure-dependent tube expansion along the transmission line and fluid inertia have not yet been fully incorporated. Extending the model to include these effects is expected to further improve estimation accuracy.
\subsubsection{Friction modeling} Explicit friction modeling (e.g., seal friction) was omitted in this study.
Although the volumetric loss model correctly accounts for pressure regardless of its source (load or friction), unmodeled friction can cause stick-slip phenomena and steady-state errors.
Integrating advanced friction compensation or low-friction hardware, such as rolling diaphragm cylinders~\cite{whitney2014low}, could further enhance positioning precision.
In summary, this baseline implementation highlights the inherent effectiveness of the SHAPE method in enabling feedback control without actuator-side sensors.
While the integration of sophisticated dynamic compensation or advanced control strategies could further enhance performance, 
the current framework provides a baseline for achieving accurate final positioning in remote robotic operations within extreme environments.

\subsubsection{Tube characteristic change model}

Temperature drift is a representative source of tube characteristic change because the present model assumes a constant Young's modulus $E$.
As \eqref{eq_x2_hat} includes tube expansion governed by $E$, the sensitivity of $\hat{x}_2$ to temperature can be evaluated by combining its sensitivity to $E$ with $dE/dT$.
The nylon PA12 showed a decrease in $E$ of approximately 12\% from 23 to 37°C, or approximately 1\% per 1°C near room temperature~\cite{amstutz2021temperature}; in the present model, a 1\% change from 287~MPa changed the estimated position by up to approximately 1~mm.
Other factors, such as water absorption of nylon tubes and radiation-induced degradation, may also change $E$ during long-term deployment in harsh environments.
Thus, under current limitations, in-situ parameter re-identification immediately prior to tasks (Section \ref{In-situ reidentification}) is desirable.
However, this study does not include an independent long-term validation dataset for tube degradation or radiation-induced hardening; evaluating reliability under realistic aging and harsh-environment exposure remains an important future challenge.

Furthermore, creep can cause pressure-deformation hysteresis. Although permanent deformation was absent between 0 and 3 MPa in separate trials, a hysteresis loop was observed. Incorporating temperature dependence and rate-dependent creep modeling remains a future challenge to further enhance estimation accuracy.

\section{Conclusion}

This study presented the SHAPE framework, which enables the transmission of driving power to the actuator and position information from it, respectively, through a long and thin water-filled tube. A primary challenge in such long-distance hydraulic systems is the low rigidity of the transmission path, which necessitates a physics-based model to compensate for volumetric losses.

By explicitly modeling pressure-dependent tube expansion and fluid pseudo-compressibility, a fundamental infrastructure that enables the application of a closed-loop feedback control framework to sensorless remote actuators was established.

The experimental validation demonstrated that the estimated position could be effectively used as a feedback signal for stable position control under varying load disturbances and the inherent low mechanical impedance of the system.

Furthermore, the proposed in-situ re-identification strategy aims to maintain estimation accuracy during long-term missions by addressing gradual environmental drifts.

Collectively, these contributions advance the feasibility of reliable remote robotics in harsh environments by eliminating actuator-side electronics that can become points of failure, while supporting accurate and safe kinematic control.

\appendices
\section{Two-Segment Pressure Propagation Model}
\label{app:two_segment_pressure_propagation}

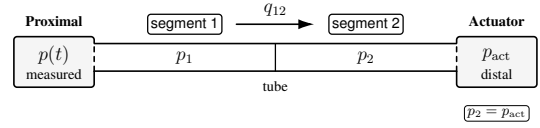
\begin{figure}[t]
    \centering
    
    \resizebox{0.8\linewidth}{!}{
    \begin{tikzpicture}[>=Latex, line join=round, line cap=round, font=\sffamily\small]

        \draw[line width=1.0pt, rounded corners=2pt, fill=gray!8] (0.4,1.35) rectangle (2.15,2.45);
        \node[font=\bfseries\small] at (1.275,2.78) {Proximal};
        \node[font=\large] at (1.275,2.05) {$p(t)$};
        \node[font=\small] at (1.275,1.63) {measured};

        \draw[line width=1.0pt, rounded corners=2pt, fill=gray!8] (10.05,1.35) rectangle (11.8,2.45);
        \node[font=\bfseries\small] at (10.925,2.78) {Actuator};
        \node[font=\large] at (10.925,2.05) {$p_{\mathrm{act}}$};
        \node[font=\small] at (10.925,1.63) {distal};

        \fill[white] (2.12,1.70) rectangle (10.08,2.30);
        \draw[line width=1.0pt] (2.15,1.70) -- (10.05,1.70); 
        \draw[line width=1.0pt] (2.15,2.30) -- (10.05,2.30); 
        \draw[line width=0.8pt] (6.1,1.70) -- (6.1,2.30);

        \draw[line width=1.0pt, dashed] (2.15,1.70) -- (2.15,2.30); 
        \draw[line width=1.0pt, dashed] (10.05,1.70) -- (10.05,2.30);

        \node[draw, rounded corners=2pt, line width=0.8pt, inner sep=2pt] at (4.125,2.72) {segment 1};
        \node[draw, rounded corners=2pt, line width=0.8pt, inner sep=2pt] at (8.075,2.72) {segment 2};
        \node[font=\large] at (4.125,2.00) {$p_1$};
        \node[font=\large] at (8.075,2.00) {$p_2$};

        \draw[-{Latex[length=3.5mm,width=2.2mm]}, line width=1.0pt] (5.25,2.72) -- (6.95,2.72);
        \node[font=\large, fill=white, inner sep=1pt] at (6.1,3.04) {$q_{12}$};

        \node[draw, rounded corners=2pt, align=center, font=\small, inner sep=2pt] at (10.925,0.78) {$p_2=p_\mathrm{act}$};

        \node[font=\small] at (6.1,1.38) {tube};
    \end{tikzpicture}
    }
    \caption{Two-segment tube model.}
    \label{fig:two_segment_tube_model}
\end{figure}

A minimal extension to address transient pressure-propagation effects is the two-segment model illustrated in Fig.~\ref{fig:two_segment_tube_model}. If $q_{12}$ is the flow rate between the two segments, the following formulation can be introduced:
\begin{equation}
q_{12}(t) = \mathcal{G}(p_1(t), p_2(t)) \label{eq_q12_operator}
\end{equation}
where $\mathcal{G}$ is a transmission dynamics operator that maps the segment pressures to the inter-segment flow. 
This formulation represents the tube dynamics as a two-state pressure propagation model rather than a purely quasi-static relation.

From the conservation of volume, the segment pressures are governed by,
\begin{equation}
\begin{aligned}
C_1(p_1)\dot{p}_1 &= A_1\dot{x}_1 - q_{12} \\
C_2(p_2)\dot{p}_2 &= q_{12} - A_2\dot{x}_2,
\end{aligned}
\label{eq:tube_governing} 
\end{equation}
where $x_1$ and $x_2$ represent the displacement of the DFMC and actuator, respectively.
$C_1$ and $C_2$ represent the effective compliance of the HST; while these were constants in previous studies, future work will introduce pressure dependence.

By using these models to estimate the pressure on the actuator side, it is expected that the accuracy of actuator position estimation will be improved.
For the simulation in Section~\ref{experimental_condition}, the pressure $\hat p_2$ was obtained using a Kalman filter with a two-segment model, where \eqref{eq_q12_operator} was linearized into the form of \eqref{eq:q_12}. In the Kalman filter, \eqref{eq:tube_governing} was used as the state equation, and $x_1$ and $p=p_1$ were the observed values.
\begin{align}
    q_{12}&=\frac{\pi r_\mathrm{i}^4}{8\mu l }\left(p_1-p_2\right), \label{eq:q_12}
\end{align}
where $\mu$ is the viscosity coefficient of water.

\begin{table}[t]
\centering
\footnotesize
\begin{threeparttable}
\caption{Comparison of Position Estimation Errors Across Different Pressure Information ($\mathrm{mm}$)}
\label{tab:comparison_error_from_pressure}
\begin{tabular}{llccc} 
\hline
\textbf{Item} & \textbf{Metric} & \textbf{Cond. (a)} & \textbf{Cond. (b)} & \textbf{Cond. (c)} \\ \hline
\multirow{2}{*}{$e_\mathrm{est}(p)$}  & RMSE & \phantom{0}1.414 & \phantom{0}1.495 & \phantom{0}1.664 \\
 & Max. error & \phantom{0}9.157 & \phantom{0}9.157 & 10.057 \\ \hline
\multirow{2}{*}{$e_\mathrm{est}(p,\hat p_2)$\tnote{$\dagger$1}}  & RMSE & 0.927  & 1.149 & 1.249 \\ 
 & Max. error & 6.508 & 7.146 & 7.722 \\ \hline
 \multirow{2}{*}{$e_\mathrm{est}(p,p_{act})$\tnote{$\dagger$2}}  & RMSE & 0.838 & 1.096 & 1.115 \\ 
 & Max. error & 3.937 & 4.996 & 4.742 \\ \hline 

\end{tabular}
\begin{tablenotes}[flushleft] 
    \item[$\dagger$1] $e_{est}(p,\hat p_2) = x_2 - \hat{x}_2(p,\hat p_2)$
    \item[$\dagger$2] $e_{est}(p,p_{act}) = x_2 - \hat{x}_2(p,p_\mathrm{act})$
\end{tablenotes}
\end{threeparttable}
\end{table}

\bibliographystyle{IEEEtran}

\bibliography{IEEEabrv,bibliography_after}\

@article{suzumori2020new,
  title={New robotics pioneered by fluid power},
  author={Suzumori, Koichi},
  journal={Journal of Robotics and Mechatronics},
  volume={32},
  number={5},
  pages={854--862},
  year={2020},
  publisher={Fuji Technology Press Ltd.}
}

@inproceedings{kobayashi2023discussion,
  title={Discussion of Teleoperation Support System Design Guideline for Working Robot in Radiation Narrow Environment},
  author={Kobayashi, Ryosuke and Yoshimi, Takashi},
  booktitle={2023 23rd International Conference on Control, Automation and Systems (ICCAS)},
  pages={1180--1184},
  year={2023},
  organization={IEEE}
}

@article{sharp1996radiation,
  title={Radiation tolerance of components and materials in nuclear robot applications},
  author={Sharp, Richard and Decreton, Marc},
  journal={Reliability Engineering \& System Safety},
  volume={53},
  number={3},
  pages={291--299},
  year={1996},
  publisher={Elsevier}
}

@inproceedings{Fukumoto2018,
  author    = {Fukumoto, Takuya and Ishizawa, Kouji and Okada, Satoshi and Hirano, Katsuhiko and Kurosawa, Kouichi and Murai, Yoichi},
  title     = {Development of Robots Working in High Radiation Environment for Decommissioning {Fukushima} {Daiichi} Nuclear Power Plant {(in Japanese)}},
  booktitle = {Proceedings of the National Symposium on Power and Energy Systems},
  volume    = {2018.23},
  pages     = {A112},
  year      = {2018},
  doi       = {10.1299/jsmepes.2018.23.A112}
}

@inproceedings{ganesh2004dynamics,
  title={Dynamics and control of an {MRI} compatible master-slave system with hydrostatic transmission},
  author={Ganesh, G and Gassert, Roger and Burdet, Etienne and Bleuler, Hannes},
  booktitle={IEEE International Conference on Robotics and Automation, 2004. Proceedings. ICRA'04. 2004},
  volume={2},
  pages={1288--1294 Vol.2},
  year={2004},
  organization={IEEE}
}

@inproceedings{whitney2014low,
  title={A low-friction passive fluid transmission and fluid-tendon soft actuator},
  author={Whitney, John P and Glisson, Matthew F and Brockmeyer, Eric L and Hodgins, Jessica K},
  booktitle={2014 IEEE/RSJ international conference on intelligent robots and systems},
  pages={2801--2808},
  year={2014},
  organization={IEEE}
}

@article{veronneau2018high,
  title={A high-bandwidth back-drivable hydrostatic power distribution system for exoskeletons based on magnetorheological clutches},
  author={V{\'e}ronneau, Catherine and Bigu{\'e}, Jean-Philippe Lucking and Lussier-Desbiens, Alexis and Plante, Jean-S{\'e}bastien},
  journal={IEEE Robotics and Automation Letters},
  volume={3},
  number={3},
  pages={2592--2599},
  year={2018},
  publisher={IEEE}
}

@article{guo2018compact,
  title={Compact design of a hydraulic driving robot for intraoperative {MRI}-guided bilateral stereotactic neurosurgery},
  author={Guo, Ziyan and Dong, Ziyang and Lee, Kit-Hang and Cheung, Chim Lee and Fu, Hing-Choi and Ho, Justin DL and He, Haokun and Poon, Wai-Sang and Chan, Danny Tat-Ming and Kwok, Ka-Wai},
  journal={IEEE Robotics and Automation Letters},
  volume={3},
  number={3},
  pages={2515--2522},
  year={2018},
  publisher={IEEE}
}

@inproceedings{watanabe2024fault,
  title={Fault Detection and Fault-Tolerant Control for Water Hydraulic Robots Driven by Air-Hydraulic Servo Booster},
  author={Watanabe, Yuki and Hyon, Sang-Ho},
  booktitle={2024 IEEE/SICE International Symposium on System Integration (SII)},
  pages={1399--1404},
  year={2024},
  organization={IEEE}
}

@article{simonelli2019hydrostatic,
  title={Hydrostatic actuation for remote operations in {MR} environment},
  author={Simonelli, James and Lee, Yu-Hsiu and Chen, Cheng-Wei and Li, Xinzhou and Mikaiel, Samantha and Lu, David and Wu, Holden H and Tsao, Tsu-Chin},
  journal={IEEE/ASME Transactions on Mechatronics},
  volume={25},
  number={2},
  pages={894--905},
  year={2020},
  publisher={IEEE}
}

@article{dong2019high,
  title={High-performance continuous hydraulic motor for {MR} safe robotic teleoperation},
  author={Dong, Ziyang and Guo, Ziyan and Lee, Kit-Hang and Fang, Ge and Tang, Wai Lun and Chang, Hing-Chiu and Chan, Danny Tat Ming and Kwok, Ka-Wai},
  journal={IEEE Robotics and Automation Letters},
  volume={4},
  number={2},
  pages={1964--1971},
  year={2019},
  publisher={IEEE}
}

@book{Matthiesen2018,
author = {Katharina Schmitz, Hubertus Murrenhoff},
title = {Grundlagen der Fluidtechnik},
series  = {Reihe Fluidtechnik},
volume   = {2},
publisher = {Shaker Verlag},
year = {2018},
address = {Aachen},
isbn = {9783844062465}
}

@article{hyon2021air,
  title={Air-hydraulic servo booster toward submersible water-driven robots},
  author={Hyon, Sang-Ho and Akama, Kazuto},
  journal={IEEE Robotics and Automation Letters},
  volume={6},
  number={2},
  pages={1966--1972},
  year={2021},
  publisher={IEEE}
}

@article{park2023stiffness,
  title={Stiffness-Switchable Hydrostatic Transmission Toward Safe Physical Human--Robot Interaction},
  author={Park, Sungbin and Park, Kyungseo and Shin, Wonseok and Kim, Jung},
  journal={IEEE Transactions on Human-Machine Systems},
  volume={53},
  number={5},
  pages={855--864},
  year={2023},
  publisher={IEEE}
}

@article{urata1998development,
  title={Development of a water hydraulic servovalve},
  author={Urata, Eizo and Miyakawa, Shimpei and Yamashina, Chishiro and Nakao, Yohichi and Usami, Yuhichi and Shinoda, Masao},
  journal={JSME International Journal Series B Fluids and Thermal Engineering},
  volume={41},
  number={2},
  pages={286--294},
  year={1998},
  publisher={The Japan Society of Mechanical Engineers}
}

@inproceedings{yoshimura2026series,
  title={Series Elastic Actuated Needle Valve and Sealing Equilibrium Flow Rate Models for Precise Position Control in Single-Acting Water-Hydraulic Actuators},
  author={Yoshimura, Shuto and Nakamura, Yuki and Noda, Tomoyuki and Nakata, Yoshihito},
  booktitle={2026 IEEE/SICE International Symposium on System Integration (SII)},
  pages={850--855},
  year={2026},
  organization={IEEE}
}

@article{nakamura2025remotehydraulic,
  title={Remote Water Hydraulic Drive Robots(in Japanese)},
  author={Nakamura, Yuki and  Noda,Tomoyuki and Nakata, Yoshihiro},
  journal={Journal of the Robotics Society of Japan},
  volume={43},
  number={3},
  pages={267--270},
  year={2025},
  publisher={The Robotics Society of Japan},
}

@book{budynas1999advanced,
  title={Advanced Strength and Applied Stress Analysis},
  author={Budynas, R.G.},
  isbn={9780071160995},
  lccn={98029448},
  series={McGraw-Hill International Editions},
  year={1999},
  publisher={WCB/McGraw-Hill}
}

@book{boresi2002advanced,
  title={Advanced Mechanics of Materials},
  author={Boresi, A.P. and Schmidt, R.J.},
  isbn={9780471438816},
  lccn={2002026738},
  year={2002},
  publisher={Wiley}
}

@book{jelali2002hydraulic,
  title={Hydraulic Servo-systems: Modelling, Identification and Control},
  author={Jelali, M. and Kroll, A.},
  isbn={9781852336929},
  lccn={2002030644},
  series={Advances in Industrial Control},
  year={2002},
  publisher={Springer London}
}

@inproceedings{boisot2008failure,
  title={Failure of polyamide 11 using a damage finite elements model},
  author={Boisot, Guillaume and Fond, C and Hochstetter, Gilles and Laiarinandrasana, Lucien},
  booktitle={ECF 17},
  pages={1554--1561},
  year={2008}
}

@misc{lirias467000,
issn = {0039-2103},
journal = {Strain },
language = {eng},
number = {2},
publisher = {Wiley-Blackwell Publishing, Inc.},
title = {Variability, Heterogeneity and Anisotropy in the quasi-static response of laser sintered PA-12 components},
volume = {53},
year = {2017-04-01},
author = {Faes, Matthias and Wang, Yueqi and Lava, Pascal and Moens, David},
}

@article{Bolignari2020,
author = {Bolignari, Marco and Fontana, Marco},
year = {2020},
month = {03},
pages = {},
title = {Design and experimental characterization of a high performance hydrostatic transmission for robot actuation},
volume = {55},
journal = {Meccanica}
}

@ARTICLE{10547707,
  author={Asai, Hidaka and Noda, Tomoyuki and Teramae, Tatsuya and Morimoto, Jun},
  journal={IEEE/ASME Transactions on Mechatronics}, 
  title={Modeling Inverse Airflow Dynamics Toward Fast Movement Generation Using Pneumatic Artificial Muscle With Long Air Tubes}, 
  year={2024},
  volume={29},
  number={4},
  pages={3038-3046},
  doi={10.1109/TMECH.2024.3400622}
}

@article{amstutz2021temperature,
  title={Temperature-dependent tensile properties of polyamide 12 for the use in percutaneous transluminal coronary angioplasty balloon catheters},
  author={Amstutz, C and Weisse, B and Valet, S and Haeberlin, A and Burger, J{\"u}rgen and Zurbuchen, Adrian},
  journal={Biomedical engineering online},
  volume={20},
  number={1},
  pages={110},
  year={2021},
  publisher={Springer}
}

\begin{IEEEbiography}[{\includegraphics[width=1in,height=1.25in,clip,keepaspectratio]{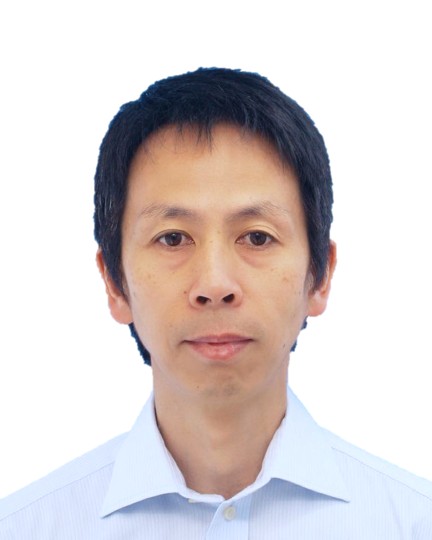}}]{Yuki Nakamura}
(M'24) received his M.E. degree in Engineering from Hiroshima University, Hiroshima, Japan, in 2004. He is currently an engineer at Chugai-Technos Corporation, Hiroshima, Japan, and is pursuing a Ph.D. degree in Robotics at the University of Electro-Communications, Tokyo, Japan. His research interests include water-hydraulic robots and remote robotic systems for harsh environments. 
\end{IEEEbiography}
\begin{IEEEbiography}[{\includegraphics[width=1in,height=1.25in,clip,keepaspectratio]{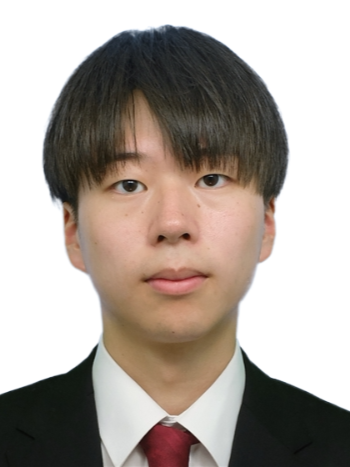}}]{Shuto Yoshimura}
(M'24) received his B.E. degree in Engineering from the University of Electro-Communications, Tokyo, Japan, in 2024. He is currently working toward the M.S. degree in Robotics at the University of Electro-Communications.
His research interests include the control of remote robots operating in harsh environments.
\end{IEEEbiography}
\begin{IEEEbiography}[{\includegraphics[width=1in,height=1.25in,clip,keepaspectratio]{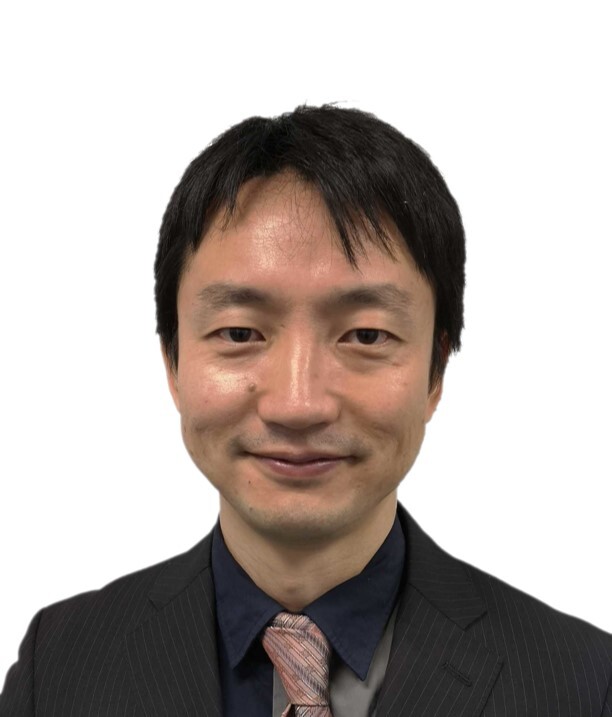}}]{Tomoyuki Noda}
(M'15) received his B.E., M.E., and Ph.D. degrees in Engineering from Osaka University, Osaka, Japan, in 2004, 2006, and 2009, respectively. He is currently with the Department of Brain Robot Interface, ATR Computational Neuroscience Laboratories, Kyoto, Japan. His research interests include brain--robot interfaces, wearable robots, and rehabilitation robotics.
\end{IEEEbiography}
\begin{IEEEbiography}[{\includegraphics[width=1in,height=1.25in,clip,keepaspectratio]{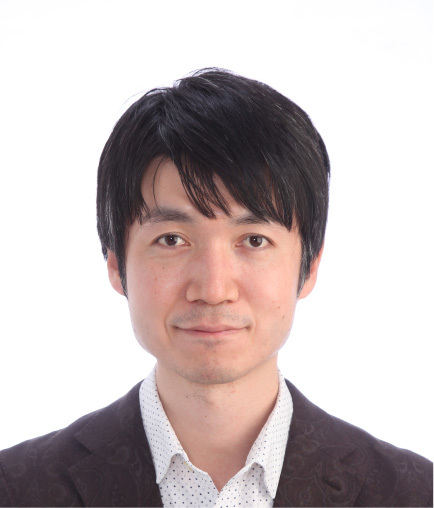}}]{Yoshihiro Nakata}
(M'13) received his B.E. and M.E. degrees and a Ph.D. in Engineering from Osaka University, Osaka, Japan in 2008, 2010, and 2013, respectively.
He is currently an Associate Professor in the Department of Mechanical and Intelligent Systems Engineering, Graduate School of Informatics and Engineering, University of Electro-Communications, Tokyo, Japan.
His research interests include actuator technologies and physical human-robot interactions.
\end{IEEEbiography}

\end{document}